\documentclass[11pt]{article}

\PassOptionsToPackage{dvipsnames, x11names, svgnames, table}{xcolor}

\usepackage[final]{acl2023}
\usepackage{times}
\usepackage{latexsym}
\usepackage[T1]{fontenc}
\usepackage[utf8]{inputenc}
\usepackage{inconsolata}
\usepackage{graphicx}
\usepackage{booktabs} 
\usepackage{hyperref}
\usepackage{etoolbox}
\usepackage{amsmath}
\usepackage{placeins}
\usepackage{adjustbox}
\usepackage{subcaption}
\usepackage{graphicx}
\newcommand{\startappendixtoc}{
  \begingroup
  \let\origcontentsline\contentsline
  \renewcommand{\contentsline}[3]{%
    \origcontentsline{##1}{##2}{##3}%
  }%
}

\makeatletter
\newcommand{\printappendixtoc}{%
  \section*{Appendix Contents}
  \@starttoc{apx}
}
\makeatother

\usepackage{amsmath}
\usepackage{amssymb}
\usepackage{pifont} 

\usepackage{multirow}
\usepackage{array}
\usepackage{makecell}
\usepackage{tabularx}
\usepackage{cellspace}
\usepackage{float}
\usepackage{rotating}
\usepackage{caption}
\usepackage{subcaption}
\usepackage{algorithm}
\usepackage{algpseudocode}
\usepackage{tikz}
\usepackage{pgfplots}
\pgfplotsset{compat=1.18}
\usepackage{tcolorbox}
\usepackage{fontawesome}
\usepackage{twemojis}

\usepackage{tipa}
\usepackage{expex}
\usepackage{linguex}

\usepackage{cleveref}

\definecolor{aclpurple}{HTML}{800080}

\newcolumntype{Y}{>{\raggedright\arraybackslash}X}

\newif\ifshowemoji

\showemojifalse  

\usepackage{graphicx}
\usepackage{titling} 
\usepackage{multirow}
\usepackage{adjustbox}
\usepackage{graphicx}
\usepackage{accsupp}
\usepackage{pdfcomment}
\newcommand{\beetle}{Beetle} 
\newcommand{\beetlelm}{\textsc{\textbf{BeetleLM}}}
\usepackage{tabularray}
\usepackage{setspace}

\newif\ifshowemoji
\showemojitrue   

\title{Beetle: A Bilingual Model Suite \\ 
for Modelling Second-Language Processing 
}

\author{
 \textbf{Suchir Salhan\,\raisebox{-0.2em}{\includegraphics[height=1em]{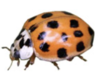}}},
 \textbf{Catherine Arnett\,\raisebox{-0.2em}{\includegraphics[height=1em]{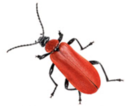}}},
 \textbf{James A. Michaelov\,\raisebox{-0.2em}{\includegraphics[height=1em]{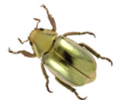}}} \&
 \textbf{Paula Buttery\,\raisebox{-0.2em}{\includegraphics[height=1em]{suchir_ladybug.png}}}
\\
\\
 \raisebox{-0.2em}{\includegraphics[height=1em]{suchir_ladybug.png}} University of Cambridge \quad
 \raisebox{-0.2em}{\includegraphics[height=1em]{catherine_firefly.png}} EleutherAI \quad
 \raisebox{-0.2em}{\includegraphics[height=1em]{james_scarab.png}} MIT
\\
 \small{
   \textbf{Correspondence:} \href{mailto:sas245@cam.ac.uk}{sas245@cam.ac.uk}
 }
}

\date{} 
\begin{document}
\maketitle
\begin{abstract}

Bilingual language models (LMs) offer a controlled setting for studying how training conditions shape second-language (L2) behaviour, but prior work typically varies exposure structure, scale, and architecture at once, making it difficult to attribute effects to any single factor. We introduce \beetle{}, a controlled language model pretraining framework in which tokeniser, target language, training budget, and exposure structure are each independently manipulable, enabling systematic and comparable experimentation of training conditions. Using \beetle{}, we train and release 285 bilingual and 45 monolingual open-source LMs with rich checkpoints across a range of exposure schedules, data scales and first languages (L1s) to study multilingual pretraining and computational modelling of bilingualism and second language learning.  Evaluating models on human bilingual and second language reading-time prediction and grammaticality judgement tasks, we find that staged and temporally structured curricula consistently improve alignment with language learner reading time compared to balanced bilingual training, with the largest gains at smaller data scales and for typologically closer language pairs.  The \beetle{} models are well suited tools to help move computational psycholinguistics beyond its prevailing monolingual, English-centric focus toward models of human bilingual processing, to study cross-lingual learning dynamics, while supporting community-based development of controlled model families.

\end{abstract}
\vspace{-0.6em}
\begin{center}
    \raisebox{-0.45em}{\includegraphics[height=1.5em]{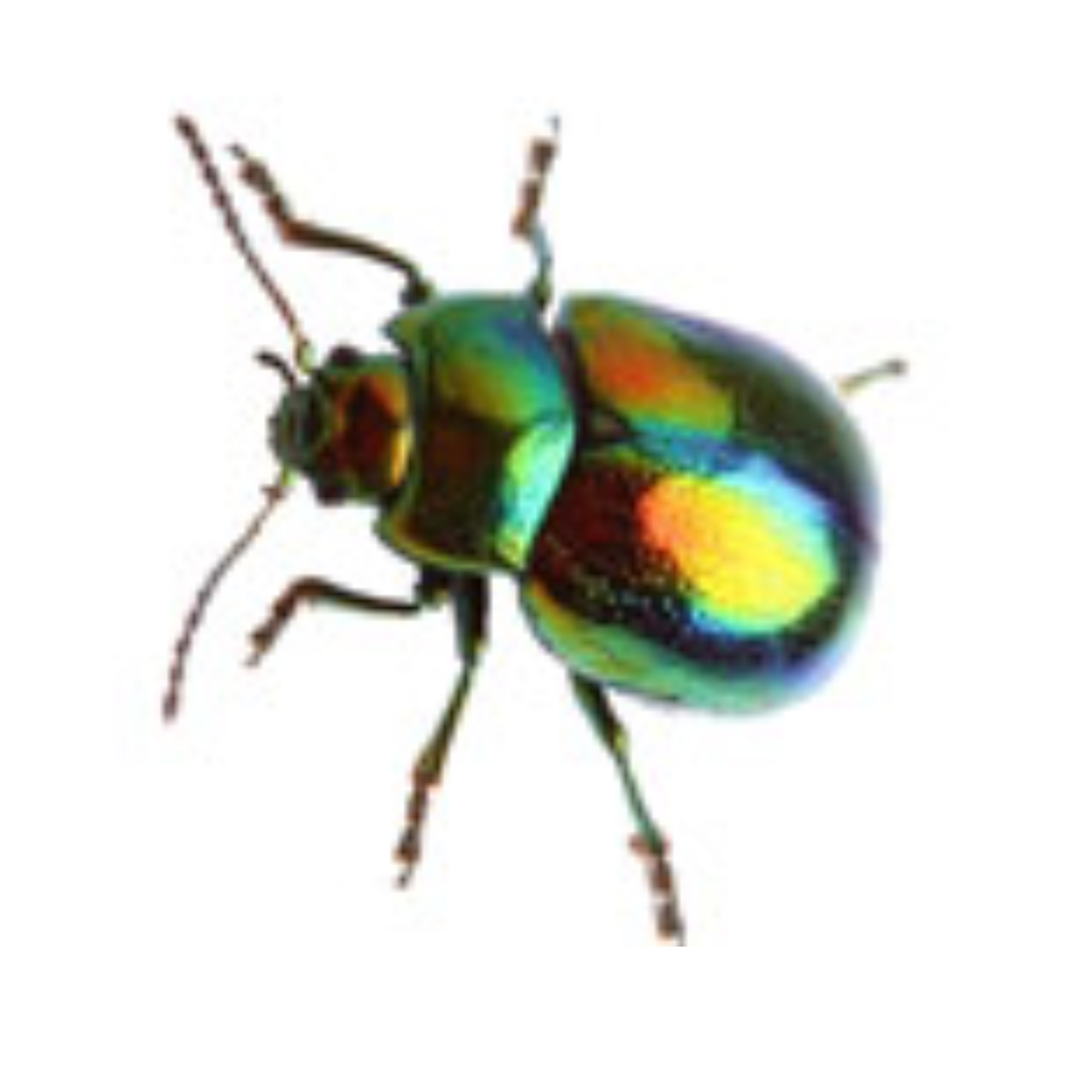}}\,
    \href{https://beetlelm.github.io/}{\texttt{beetlelm.github.io}}
\end{center}
\vspace{-0.5em}

 \section{Introduction}

\begin{figure}[t!]
    \centering
    \includegraphics[width=\linewidth]{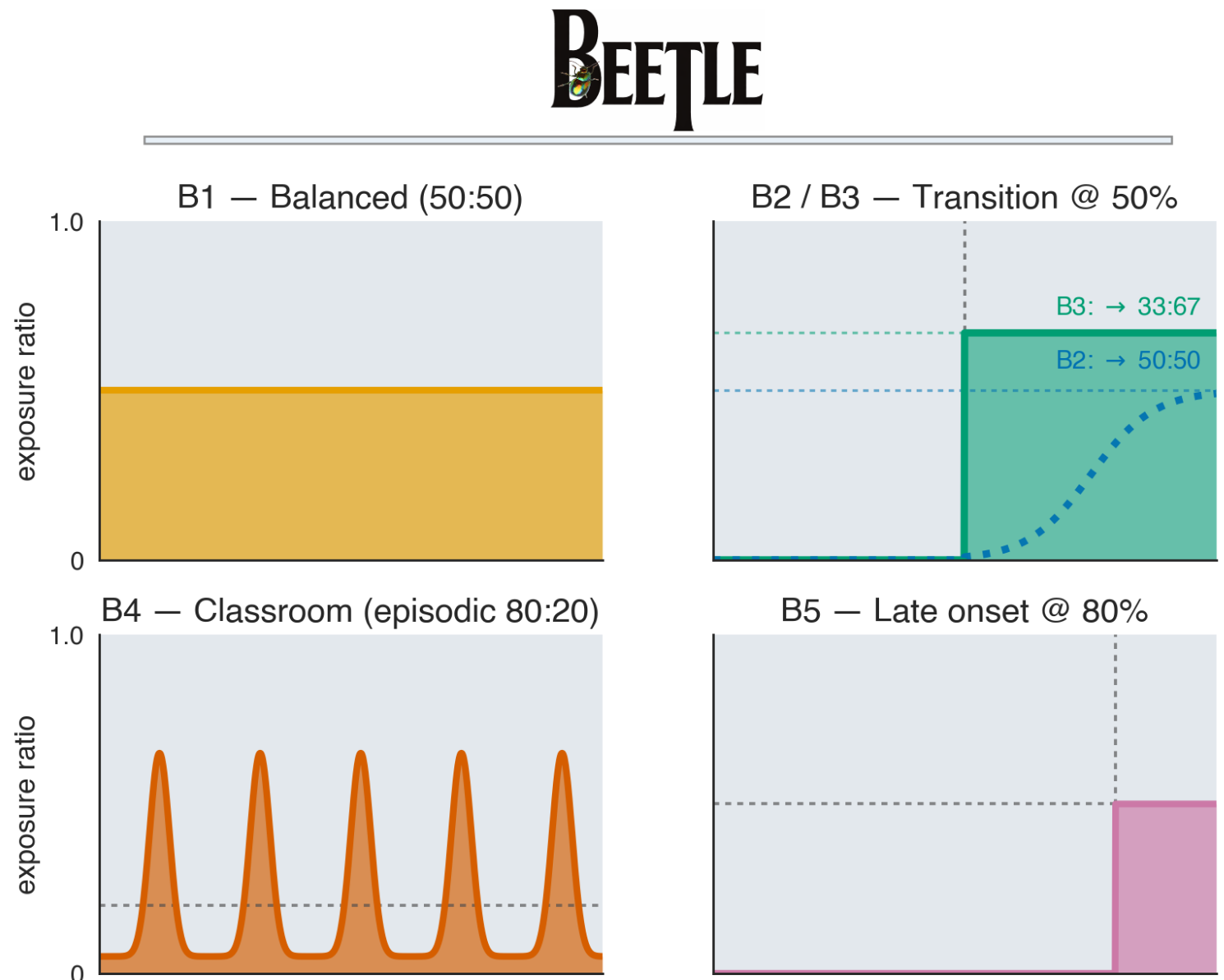}
    \caption{\beetle{} is an open-source language model pretraining framework. We use it to study bilingual and multilingual language acquisition in language models under tightly controlled experimental conditions. Each facet shows different structured exposure curriculum (\texttt{B1--B5}) used to pretrain bilingual models that vary in the relative per-language data distribution across training steps.}
    \label{fig:beetle_splash}
\end{figure}

In contrast to multilingual models, bilingual and second-language (L2) language models (LMs) are models trained on exactly two languages, either jointly or in sequence. These models are a useful middle ground between monolingual and massively multilingual systems for studying multilingual language model pretraining. They also allow us to study bi- and multilingual language acquisition, representations, interpretability, and learner simulation \citep{frank2021toward,yadavalli-etal-2023-slabert, oba-etal-2023-second, aoyama2024modeling, constantinescu2025investigating, arnett-etal-2025-acquisition}. However, important questions about the dynamics of bilingual training remain open. When does cross-linguistic transfer emerge during pretraining, how does transfer depend on typological similarity, relative timing and distribution of exposure to each language, and how does joint training compare with learning each language independently \citep{chang-etal-2022-geometry,blevins2022analyzing,wang2024probing,rajaee2024analyzing,korner-etal-2026-meanings, zhou2026crosslingualalignmentjointtraining}? Existing bilingual models provide useful evidence for transfer, but are often constrained by the available training setups. The only open bilingual family with checkpoints, B-GPT \citep{arnett-etal-2025-acquisition}, was developed for only a limited set of language pairs and does not provide fine-grained checkpoints over the early stages of training, making it difficult to distinguish effects that emerge during initial acquisition from those that arise later in training. Moreover, the absence of matched monolingual baselines makes it difficult to determine whether apparent bilingual benefits reflect genuine transfer between languages or simply differences in the amount and composition of training data. These limitations motivate controlled bilingual pretraining experiments that vary exposure structure while tracking learning from the earliest stages and comparing bilingual models against matched monolingual controls.

Despite the fact that most people speak at least two languages \citep{grosjean2021life}, most cognitive science research focuses on monolingual language processing. In computational cognitive science, one limitation has been the lack of available bilingual language models. Additionally, the studies that have been done differ significantly in training scale, exposure schedule, language pair, continual-learning mechanism, and evaluation protocol. As a result, it is hard to attribute behaviour of a model to any single training decision. 

To address this, we introduce \beetle{}, a modular LM pretraining and analysis framework that controls data exposure, alongside the usual choices of data quantity, tokeniser, and architecture (\S\ref{sec:framework}), to compare and investigate these confounding factors. We unify previously proposed methods in our training framework.
and use our models to investigate whether acquisition-inspired language exposure curricula during bilingual LM pretraining leads to better alignment with metrics of human language processing.

Using our training framework, we train a suite of \textbf{285 bilingual models} spanning 21 L1s, holding  architecture, model size, tokenizer, and overall language mixture fixed. We vary two factors: the L1 (English is always the L2) and the order in which the two  languages are presented during pretraining. We train bilingual models with 5  exposure conditions at three FineWeb dataset scales (100M, 2B and 24B tokens), plus a matched human-scale 100M condition.  We additionally provide \textbf{45 monolingual baselines covering 20 languages}. Every \beetle{} model is released with its full checkpoint trajectory for studying cross-lingual learning dynamics.

We find that structured exposure improves alignment with human eye-tracking reading times on Multilingual Eye-movement Corpus (MECO) L2 data \citep{kuperman2023text,kuperman2025new} and that ordered curricula tend to outperform fully interleaved training, suggesting that gradual language introduction improves alignment with psycholinguistic data. These effects are not explained by differences in overall proficiency or language balance, and remain consistent across controlled ablations. We additionally evaluate larger massively multilingual models, showing that carefully designed bilingual curricula can match or exceed L2 reading-time alignment under substantially smaller parameter budgets.  These results demonstrate that the point during training when languages are introduced is a key determinant of how closely language models reproduce human second-language reading behaviour. 

Beyond reading time, we further probe bilingual processing through cross-lingual structural priming (\S\ref{sec:structural-priming}) and CEFR-stratified L2 learner-error discrimination on BLiSS (\S\ref{sec:discrimination}), and find that curriculum effects do not transfer uniformly across these tasks: balanced exposure is often competitive or better on grammaticality and error discrimination even where staged curricula best predict reading times.

The primary contributions of this paper are 1) a flexible training framework to enable controlled multilingual pretraining with learning dynamics 2) a suite of bilingual LMs 3) and modelling efforts for bilingual data on two evaluation domains, which can inform future bilingual cognitive modelling work.

\section{Background}\label{sec:related}

\paragraph{Bilingual models.}  Bilingual and second-language language models (L2LMs) offer a controlled middle ground between monolingual and massively multilingual models, and have been used to study second-language acquisition through measures such as reading times and structural priming \citep{matusevych-etal-2013-computational, yadavalli-etal-2023-slabert, oba-etal-2023-second, aoyama2024modeling, arnett-etal-2025-acquisition, constantinescu2025investigating, feng2026bilingual}. But there is no single recipe for creating an L2LM: prior work varies in what languages are seen, when they are introduced, and how strongly earlier knowledge is preserved. Some approaches train both languages jointly, while others introduce the L2 after L1 pretraining, using continual-learning methods such as elastic weight consolidation (EWC) to limit forgetting \citep{constantinescu2025investigating}. Others manipulate the timing and distribution of L1 and L2 exposure directly, as in the structured exposure regimes of B-GPT \citep{arnett-etal-2025-acquisition}. This heterogeneity makes it difficult to disentangle effects of language pair, model scale, and exposure history. Our framework adopts this structured-exposure approach while systematically controlling these factors, enabling reproducible comparisons across bilingual training curricula.

\paragraph{Modelling human language processing.} Within psycholinguistics, LM surprisal is often used to model processing difficulty. Surprisal is incorporated as a regression predictor alongside baseline lexical covariates, and evaluated via the improvement in log-likelihood over this baseline ($\Delta\log L$; \citealp{smith2013effect,wilcox2020predictive,oh2022comparison,de-varda-marelli-2023-scaling}), making psychometric predictive power dependent on how well a model’s probability distribution aligns with human reading behaviour rather than on raw model quality alone. However, while LM–human alignment has been studied extensively in L1 settings, bilingual and L2 reading-time modelling remains comparatively under-represented \citep{frank2014modelling,de-varda-marelli-2022-effects, aoyama2024modeling}, and this gap is especially pronounced in controlled large-scale sweeps of training conditions. In this paper, we hope to help address these limitations in the literature.

\section{The \beetle{} Training Framework}\label{sec:framework}

\beetle{} is a language model pretraining framework that  allows controlled manipulation of pretraining conditions. We apply \beetle{} for controlled pretraining of variables that prior L2LM work has manipulated separately: L2 onset timing, L2 exposure ratio, and continual-learning regularisation. Language exposure in bilingual pretraining is modelled as changing continuously over the course of training, making it possible to represent gradual or abrupt transitions between languages, periodic bursts of exposure, and clustered context-dependent input.

\paragraph{Architecture.} The default backbone, \textbf{PicoDecoder} \citep{diehl-martinez-etal-2025-pico}, is a 125M-parameter LLaMA-style causal decoder \citep{touvron2023llama}: 14 layers, $d_{\mathrm{model}}=768$, 12 attention heads with one KV head (grouped-query attention), Rotary Position Embeddings, SwiGLU activations, RMSNorm, sequence length 512. This places \beetle{} on a contemporary LLaMA-class architecture (RoPE, SwiGLU, RMSNorm, grouped-query attention) that mirrors the components used in current open-weight LLMs, rather than the GPT-2-style decoder used in B-GPT \citep{arnett-etal-2025-acquisition} and the embedding-and-head adaptation of \citet{aoyama2024modeling}. Full training hyperparameters are listed in \autoref{tab:hyperparams}.

\beetle{} also contains modular logic for tokenization, potentially allowing investigation of the benefits of vocabulary overlap, different tokenization schemes (e.g., SentencePiece, UnigramLM, SuperBPE, BPE) and equal compression \citep{kallini-etal-2025-false}.  Our models use a BPE tokenizer based on the \texttt{Huggingface} tokenizers library, with a $50$K vocabulary. 
The \beetle{} models have tokenizers are trained with equal compression, so they have the same compression rate in both languages.
This means each model see the amount of information for each language in each sequence.

\paragraph{Exposure curricula.}\label{sec:exposure_curricula}
We compare five exposure curricula that manipulate how L2 input is distributed over training. Four curricula differ primarily in when L2 is introduced and how the L1:L2 mixture subsequently changes; B4 instead keeps L2 available throughout training but varies when the model sees L2 tokens, concentrating them into discrete episodes. The curricula therefore vary in L2 onset, final L1:L2 ratio, and temporal distribution, and are illustrated in Figure~\ref{fig:beetle_splash} (see \autoref{app:curricula} for details). We report the total proportion of training tokens in L2 for each curriculum, since curricula that introduce L2 at different points necessarily provide different amounts of L2 unless their later exposure compensates for the delayed onset.

\begin{itemize}
\item \textbf{B1 (Balanced):} L1 and L2 are presented in a constant 50:50 mixture throughout training. L2 is therefore available from the beginning and accounts for $50\%$ of all training tokens.

\item \textbf{B2 (Simultaneous):} Models are trained on L1 only, then halfway through training, L2 is gradually introduced using a sigmoid transition to a 50:50 mixture for the rest of training. L2 comprises \textbf{$25\%$} of all training tokens.

\item \textbf{B3 (Sequential):} Training begins with L1 only. At halfway, L2 is gradually introduced using a sigmoid transition, eventually reaching an L2-dominant 33:67 L1:L2 mixture. L2 accounts for \textbf{ $33.5\%$} of all training tokens.

\item \textbf{B4 (Classroom / episodic):} Unlike B2, B3, and B5, L2 is available throughout training rather than being introduced at a specific onset. The overall L1:L2 ratio is 80:20, but L2 tokens are temporally clustered into intermittent, high-variance episodes rather than being evenly distributed. Thus, B4 tests the effect of \emph{when L2 tokens occur within training} while keeping L2 available from the outset. L2 is $20\%$ of training tokens.

\item \textbf{B5 (Late):} Training begins with L1 only, with L2 introduced at $80\%$ of training. A sigmoid transition then gradually increases L2 exposure to a 50:50 L1:L2 mixture for the remainder of training. L2 tokens are \textbf{$10\%$}.
\end{itemize}

\section{Models \& Training}\label{sec:models-training}

We release \textbf{285 bilingual and 45 monolingual \beetle{} open-source language models} spanning \textbf{21 L1s}, with approximately 30 checkpoints per model.

\begin{figure*}[t]
    \centering
    \includegraphics[width=\linewidth]{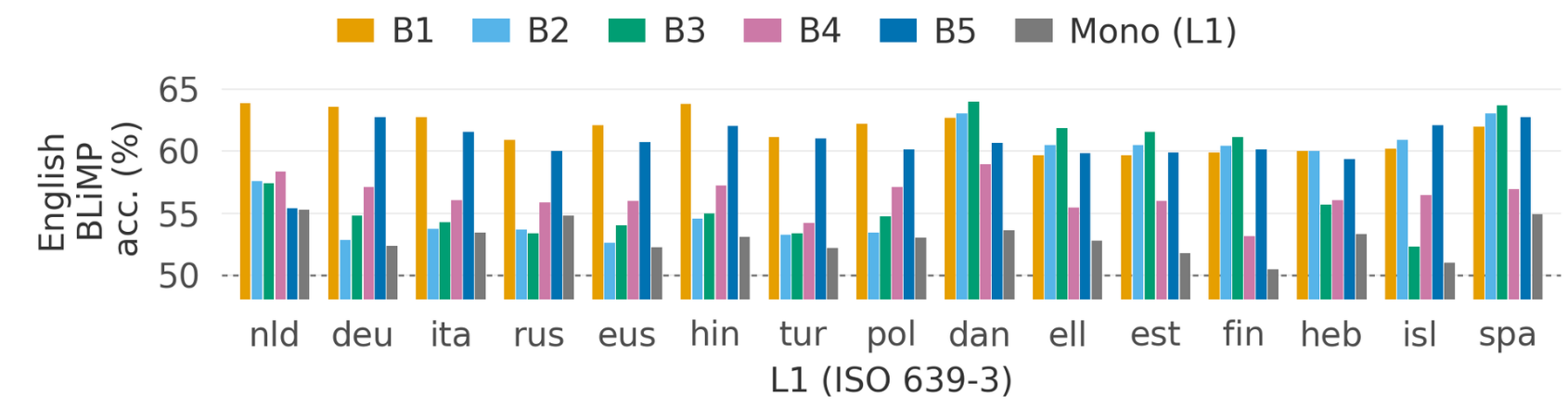}
    \caption{\beetle{} English BLiMP Accuracy for 100M FineWeb models across 15 L1s (Dutch, German, Italian, Russian, Basque, Hindi, Turkish, Polish, Danish, Greek, Estonian, Finnish, Icelandic and Spanish) and 5 exposure curricula. Full Monolingual and Multi-BLiMP accuracies for the \beetle{} models are in \autoref{blimp}.}
    \label{fig:blimp-main}
\end{figure*}

Our experiments hold a fixed architecture, tokeniser, and data source design, as independent controls to measure the effect of L1, exposure curricula (\texttt{B1 - B5})  and data scale (100M, 2B, 24B tokens) on L2 performance.  All models described in this paper have English as their L2, due to the availability of L2 reading time data (Section \ref{sec:reading-time}) and CEFR-stratified morphosyntactic and pedagogical tasks (Section \ref{sec:discrimination}).

\paragraph{Beetle-Data.} The human-scale models are trained on 100M tokens from BabyBabelLM \citep{jumelet2025babybabellmmultilingualbenchmarkdevelopmentally}; the web-scale models (at 100M, 2B, 24B tokens) use FineWeb-2 \citep{penedo2025fineweb2} for L1 and FineWeb-Edu for English.  We implement a data decontamination pipeline, removed via 13-gram overlap detection following \citet{brown2020gpt3}, for our 19 evaluation datasets (full list in \autoref{sec:beetle-data}). We release the full data tokenized in the order seen by the model for replicability and studying data memorisation. 

\paragraph{L1s.}   \beetle{} models cover a typologically diverse set of L1s, with coverage continually expanding to support broader psycholinguistic and cross-lingual modelling. At 100M FineWeb, our experiments report results on Danish, German, Greek, Estonian, Basque, Filipino, Finnish, Hebrew, Hindi, Icelandic, Italian, Japanese, Korean, Dutch, Polish, Russian, Spanish, Turkish, and Chinese. At larger data scales, we focus on 9 L1s (German, Hindi, Italian, Dutch, Polish, Russian, Spanish, Turkish, Chinese). For BabyBabelLM, we report 100M-token experiments for German, Chinese, and Dutch.

\paragraph{Checkpointing}  Checkpoints are saved according to a $\log_2$ schedule following Pythia \citep{biderman2023pythia}, which re-starts at the beginning of each new exposure phase. This  enables granular investigation of learning dynamic near phase boundaries, but with more sparse checkpointing during less critical periods of training. For each model, we release $\approx30$ checkpoints. Following \citet{diehl-martinez-etal-2025-pico}, checkpoints save model gradients, activations and circuits, facilitating detailed interpretability analyses on circuit formation in bilingual LM pretraining \citep{inaba-etal-2025-bilingual}. 
To the best of our knowledge, our models will be the first with checkpoints available for several of these languages, enabling more linguistically diverse training dynamics research. 

\paragraph{Baselines.} We compare our models against two types of baselines.  \textbf{Monolingual baselines} are matched-architecture \beetle{} models pretrained on a single language for German, Dutch and Chinese across all scales and Russian, Italian, Turkish and Basque at 100M/2B scale.  

\textbf{Massively Multilingual LLMs}: we also compare \beetle{} models to massively multilingual base models, specifically,
Apertus 8B \citep{swiss-ai-2025-apertus}, XGLM 4.5B \citep{lin-etal-2022-shot}, Llama 3.1 1B \citep{grattafiori2024llama}, Gemma 3 270M \citep{gemmateam2025gemma3technicalreport}, and Qwen 3 0.6B \citep{yang2025qwen3}. Full baseline model details are given in \autoref{tab:baselines}.

\subsection{Analysing Curriculum-Induced Learning Dynamics}\label{sec:sanity-checks}

\begin{figure*}[ht!]
    \centering

    \includegraphics[width=0.9\linewidth]{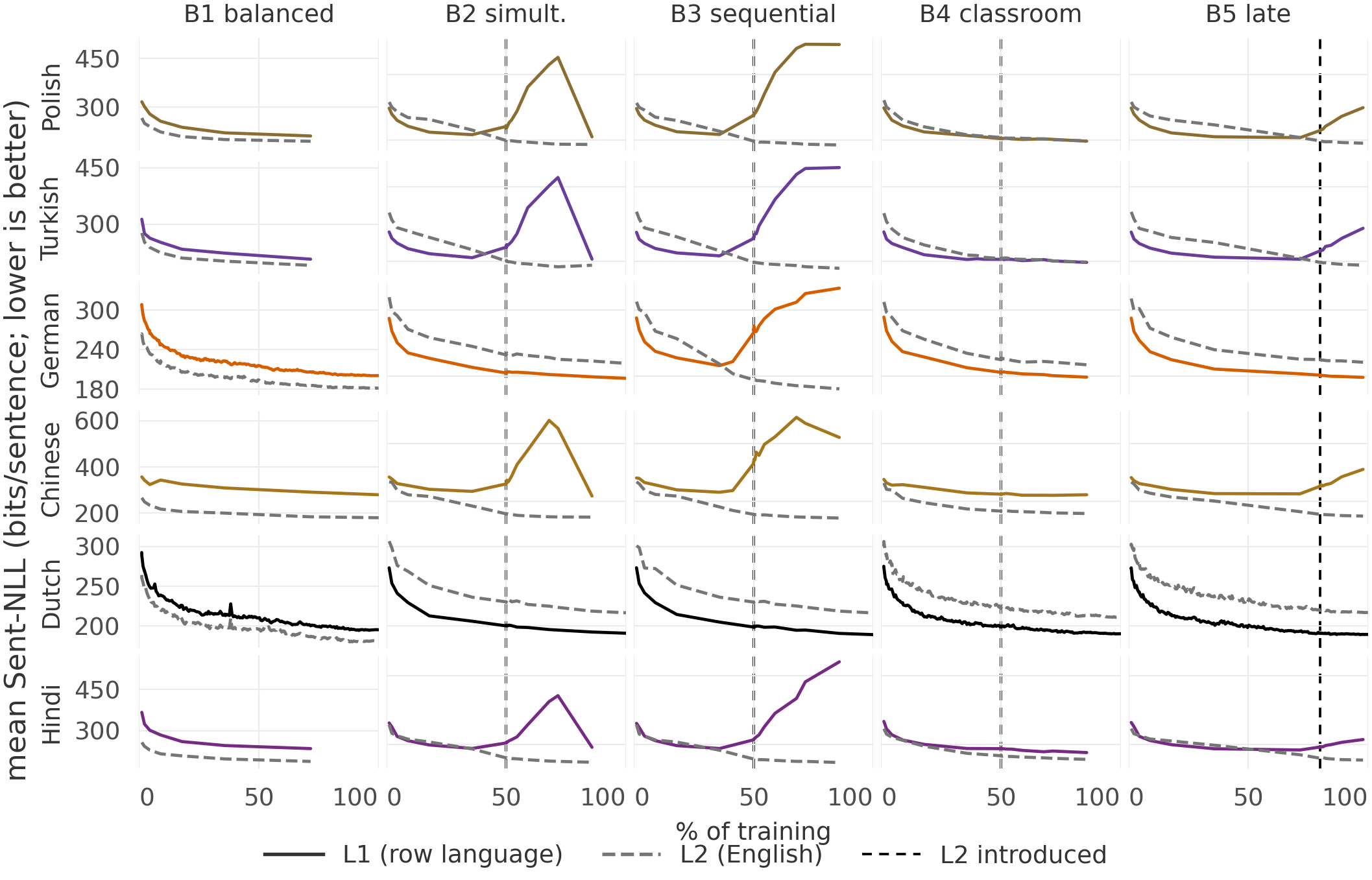}

    \caption{\textbf{FLORES Sentence NLL.} Each facet is one model; curves report mean sentence-level NLL in bits on 1,012 parallel FLORES-200 devtest sentences (lower is better), with solid lines for L1 and dashed lines for English. The vertical dashed line shows L2 English introduction.}
    \label{fig:flores_curves}

\end{figure*}

Before comparing our models with human behavioural data and metrics of bilingual language, we visualize the training dynamics (Figure~\ref{fig:flores_curves}) and test grammatical learning (Table~\ref{tab:beetle_main}) in order to validate that the exposure curricula demonstrate expected patterns. In order to avoid confounds from tokenizers or writing systems \citep{yang2026applesapplescomparablecrosslingual}, we calculate the sentence-level negative log-likelihood (Sent-NLL) for each parallel item in FLORES-200 and calculate the mean. This provides a comparable metric of model performance across languages and across models.

\paragraph{Cross-Lingual Transfer and L1 Retention} The FLORES-200 trajectories in \autoref{fig:flores_curves} reveal a strikingly asymmetric pattern of bilingual acquisition: introducing English causes an abrupt, curriculum-aligned increase in L1 sentence-level NLL, followed by partial recovery, while English NLL consistently declines after its introduction. Observed alignment of the shock is aligned with the L2 onset and subsequently recovers: L1 NLL increases under \texttt{B2}/\texttt{B3}, compared with the later-onset \texttt{B5}. NLL changes less under \texttt{B4}, where English ultimately accounts for only $20\%$ of the mixture. Chinese and Hindi exhibit large disruptions than other languages, such as Dutch and German. 

This divergence is consistent with a competition-for-capacity account of cross-lingual interference: acquiring a newly introduced language produces an immediate cost to the language already represented by the model, with the size of the cost depending strongly on the exposure schedule \citep{whitford2015second}.

\paragraph{Mono- and Multi-BLiMP.} 
We evaluate the \beetle{} models on monolingual BLiMPs (BLiMP \citep{warstadt2020blimp} for English; BLiMP-NL \citep{suijkerbuijk2025blimp} for Dutch; ZhoBLiMP \citep{liu-etal-2025-zhoblimp}) and MultiBLiMP \citep{jumelet2025multiblimp}.

At the human-scale budget of 100M tokens, \texttt{B1} produces the strongest English grammar: a mean BLiMP score of 61.6 across 15 L1s, versus 56.3 for the strongest staged curriculum, \texttt{B4}. \texttt{B1} is also the top bilingual curriculum for 9 of 15 L1s (versus 5 for \texttt{B3} and 1 for \texttt{B5}; Figure~\ref{fig:blimp-main}). Strikingly, every bilingual curriculum beats its L1-monolingual control, whose mean score is only 52.9 and remains close to chance. The advantage is even larger on MultiBLiMP~(Eng): \texttt{B1} reaches 85.2, compared with 67.4 for the monolingual controls. The benefit is language-specific, however: on MultiBLiMP~(L1), monolingual models lead with 86.0, narrowly ahead of the best bilingual curriculum, \texttt{B2} (85.1). Thus, at 100M tokens, balanced bilingual exposure is particularly effective for acquiring English grammar, without requiring a staged curriculum. This result is not driven by the expanded evaluation: within the seven-L1 subset reported in the paper, \texttt{B1} is again the strongest curriculum for all seven languages (62.6 vs.\ 54.2 for \texttt{B2}).
Full L2 English BLiMP scores across all L1s and curricula are reported in \autoref{tab:beetle_main}

\section{Reading Time (MECO L2)}\label{sec:reading-time}
The study of language processing in multilinguals is still an emerging area in computational psycholinguistics. \beetle{} provides a way to manipulate model training curricula such that we can test hypotheses about how different patterns of learning shape model predictions; and thus may enable research into how different forms of language exposure may impact patterns of reading.

Data scale is also a relevant factor in modelling human language processing. Research suggests that improvements in models' next-word prediction capability (as measured by perplexity) only lead to a better fit to reading times up to point, after which the alignment becomes worse \citep{kuribayashi-etal-2021-lower,wilcox2020predictive,oh2022comparison,oh2024frequency}. This transition has been reported to occur at around 2B training tokens \citep{oh2023transformer}, with some variation depending on reading time measure or dataset \citep{aoyama-wilcox-2025-language,michaelov2026better}. Beyond this point, the fit to larger model surprisal degrades to a greater extent \citep{oh2023transformer,michaelov2026better}, which may explain why smaller models trained on large amounts of data often perform better than larger ones \citep{oh2022comparison,oh2024frequency}. Additionally, there is a movement to use ``human-scale'' pretraining data ($\sim$100M tokens; \citealp[]{choshen2026babylm}) that is also developmentally plausible, composed of child-oriented data \citep{jumelet2025babybabellmmultilingualbenchmarkdevelopmentally}. We compare both of these data scales against near-Chinchilla-optimal \citep{hoffmann2022} web-scale data (24B tokens for 125M parameter models).

\begin{figure*}[t]
    \centering
    \includegraphics[width=\linewidth]{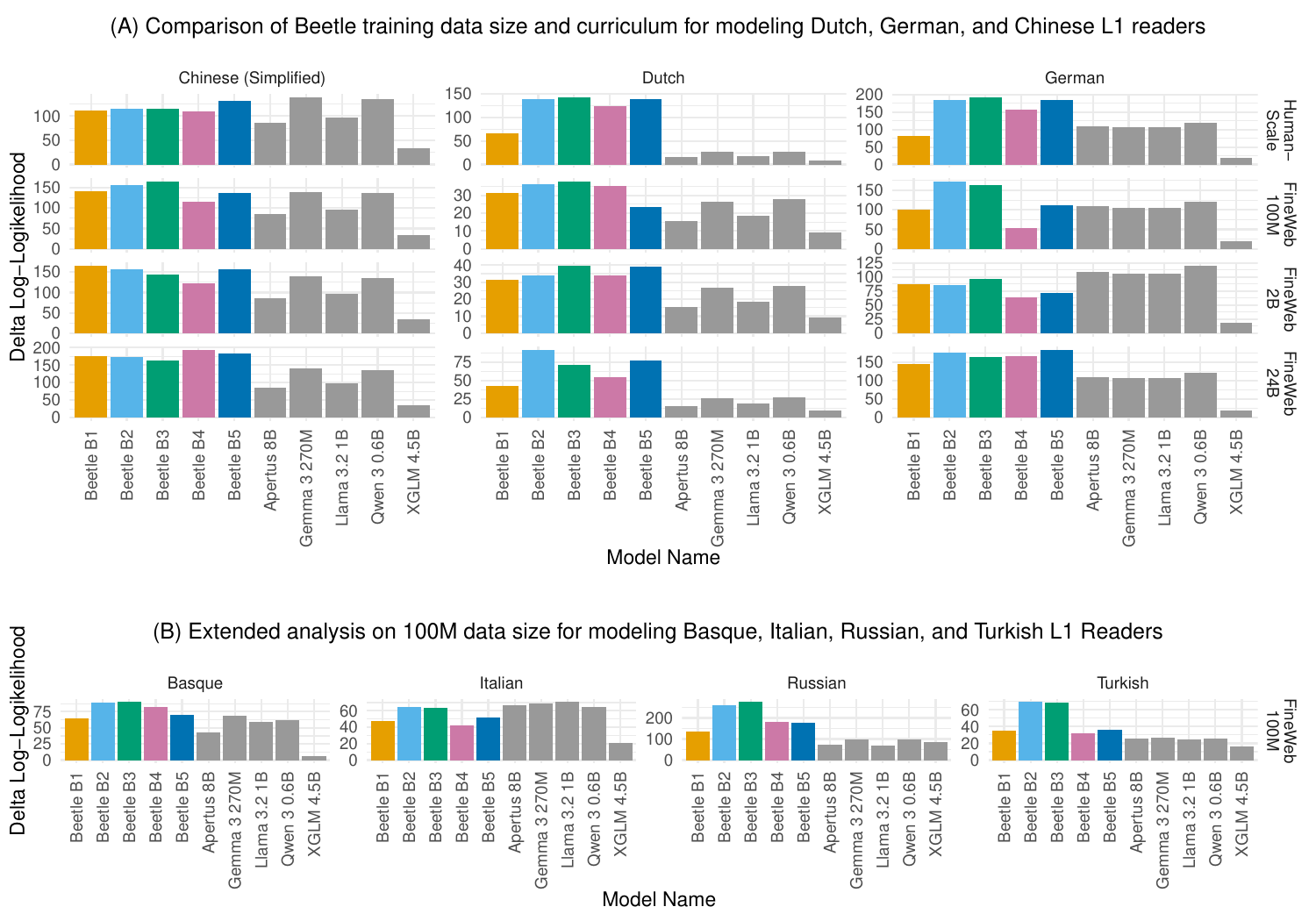}
    \caption{\textbf{Comparison of L2 Reading Time Alignment (MECO, \citep{siegelman2025wave, kuperman2025distance}) of  \beetle{ } and multilingual LLMs} : (a) \texttt{B1-B5} Dutch/Chinese/German-English \beetle{} (FineWeb 100M, 2B, 24B \citep{penedo2025fineweb2}, BabyBabelLM \citep{jumelet2025babybabellmmultilingualbenchmarkdevelopmentally}) and (b) \texttt{B1 - B3}  Basque/Italian/Russian/Turkish-English \beetle{} (100M FineWeb)}
    \label{fig:meco-main}
\end{figure*}

\paragraph{MECO eye-tracking dataset.}
We evaluate how bilingual exposure curricula affect how well \beetle's surprisal predicts human L2 reading times using the Multilingual Eye-Movement Corpus (MECO), a multilingual eye-tracking benchmark covering controlled reading in typologically diverse languages \citep{siegelman2022expanding,siegelman2025wave,kuperman2023text,kuperman2025new}. We focus on English L2 reading by Dutch, German, and Chinese L1 participants using the combined Wave~1+2 releases\footnote{\url{https://meco-read.com/}}. MECO supports controlled cross-linguistic comparisons across matched reader populations. We use \beetle{} surprisals to predict \textit{first-pass duration} (also known as\textit{ gaze duration)}, a widely-studied psycholinguistic eye-tracking measure that represents the time spent fixated on a word in the first pass before moving to another word, which shows strong surprisal-based prediction effects \citep[e.g.,][]{wilcox2020predictive}, and in the multilingual processing literature has been associated with early lexical access and initial word integration during reading \citep{dussias2010uses,luke2023exgaussian, nahatame2025relative}. We compare the \beetle{} models to other baseline LMs, selected on the basis of being massively multilingual, and thus confirmed to or likely to be trained on the languages of interest. Given the aforementioned scaling effects, we select the smallest model of each family available. Mixed-effects regressions compare models with and without surprisal predictors, with improvements measured as $\Delta \log L$(see \autoref{sec:meco-results} for full details).

\paragraph{Results} First, we consider L2 reading time for Dutch-, German-, and Chinese-L1 models. For these language pairs, we use models trained on all five curricula and four dataset sizes and compare how well they predict reading time (Figure \ref{fig:meco-main}A). 
\autoref{fig:meco-main} shows that
the MECO alignment of the \beetle{} models is comparable to, or exceeds, multilingual LLMs at a range of parameter counts. Across seven L1s in \autoref{fig:meco-main}, staged exposure curricula (\texttt{B2/B3/B5}) consistently produce stronger surprisal--reading-time alignment than balanced bilingual exposure (\texttt{B1}); an effect strongest for Russian and German/Dutch L1s. Across training scales, effects of staged exposure are strongest for Dutch and German readers and weaker for Chinese readers, suggesting that curriculum benefits interact with typological proximity to English. 

Comparing Human-Scale and FineWeb data at 100M token training budget, child-oriented training data from BabyBabelLM \citep{jumelet2025babybabellmmultilingualbenchmarkdevelopmentally} provides higher MECO alignment for Dutch and German L1 reading times, though this is not observed for Chinese L1 data. 
For Chinese, Dutch, and German, the models trained on 24B tokens show a closer fit than than at the smaller data scales, which contrasts with previous research on monolingual reading time showing a decrease in fit beyond 2B tokens \citep{oh2023transformer}. Figure \ref{fig:meco-main} shows reading time prediction for four additional L1s: Basque, Italian, Russian, and Turkish. For all language pairs except for Italian, surprisal calculated using our models better predict reading time than surprisal calculated with larger multilingual models.

\section{Bilingual Processing Effects}\label{sec:bilingual-processing}

\subsection{Structural Priming}\label{sec:structural-priming}
Previous work has argued that bilingual speakers share syntactic representations between languages, which allows for cross-linguistic syntactic priming \citep{hartsuiker2004crosslinguistic}: people are more likely to re-use the same syntactic structure in one language after hearing it in another language. It has been demonstrated that language models also exhibit crosslingual structural priming, showing that language models also share crosslingual representations \citep{michaelov-2023-structural,arnett-etal-2025-acquisition}. We replicate the genitive priming \citep{bernolet2013language} experiments in \citet{arnett-etal-2025-acquisition} with human-scale and web-scale Dutch-English \beetle{} models. Following \citet{arnett-etal-2025-acquisition}, we quantify structural priming as the difference in normalized probability of a target sentence following congruent versus incongruent primes, fitting a linear mixed-effects model for each model–language combination with prime type as a fixed effect and experimental item as a random intercept, reporting the final-checkpoint results while applying Benjamini–Hochberg false-discovery-rate correction for multiple comparisons.
We report the results in Table~\ref{tab:beetle-priming}, which show robust priming effects  across curricula for both training corpus sizes, with the exception of a marginally significant effect in the web-scale B3 model.
This shows that the \beetle{} models learn to share crosslingual representations like other bilingual models, such as the B-GPT models \citep{arnett-etal-2025-acquisition} and larger multilingual models \citep{michaelov-2023-structural}, even when our models have seen small amounts of L2 data and are trained on much smaller datasets.

\begin{table}[h!]
\centering
\small
\setlength{\tabcolsep}{4pt}
\begin{adjustbox}{max width=\columnwidth}
\begin{tabular}{lrrrr}
\toprule
 & \multicolumn{2}{c}{FineWeb} & \multicolumn{2}{c}{HumanScale} \\
\cmidrule(lr){2-3}\cmidrule(lr){4-5}
Curr. & diff & $p$ & diff & $p$ \\
\midrule
\multicolumn{5}{l}{\textit{Bernolet (genitive)}} \\
B1 & 1.051 & .0004 & 1.369 & $<$.0001 \\
B2 & 0.991 & .037  & 1.095 & $<$.0001 \\
B3 & 0.969 & .051  & 1.216 & $<$.0001 \\
B4 & 1.252 & .0008 & 0.565 & .005 \\
B5 & 1.075 & .0003 & 0.571 & $<$.0001 \\

\bottomrule
\end{tabular}
\end{adjustbox}
\caption{Cross-lingual structural priming for the 24B FineWeb and HumanScale models: surprisal difference (nats) between congruent and incongruent primes, with permutation $p$-values. }
\label{tab:beetle-priming}
\end{table}

\subsection{L2 Discrimination \& CEFR}\label{sec:discrimination}

\begin{figure}[!t]
  \centering
  \includegraphics[width=\linewidth]{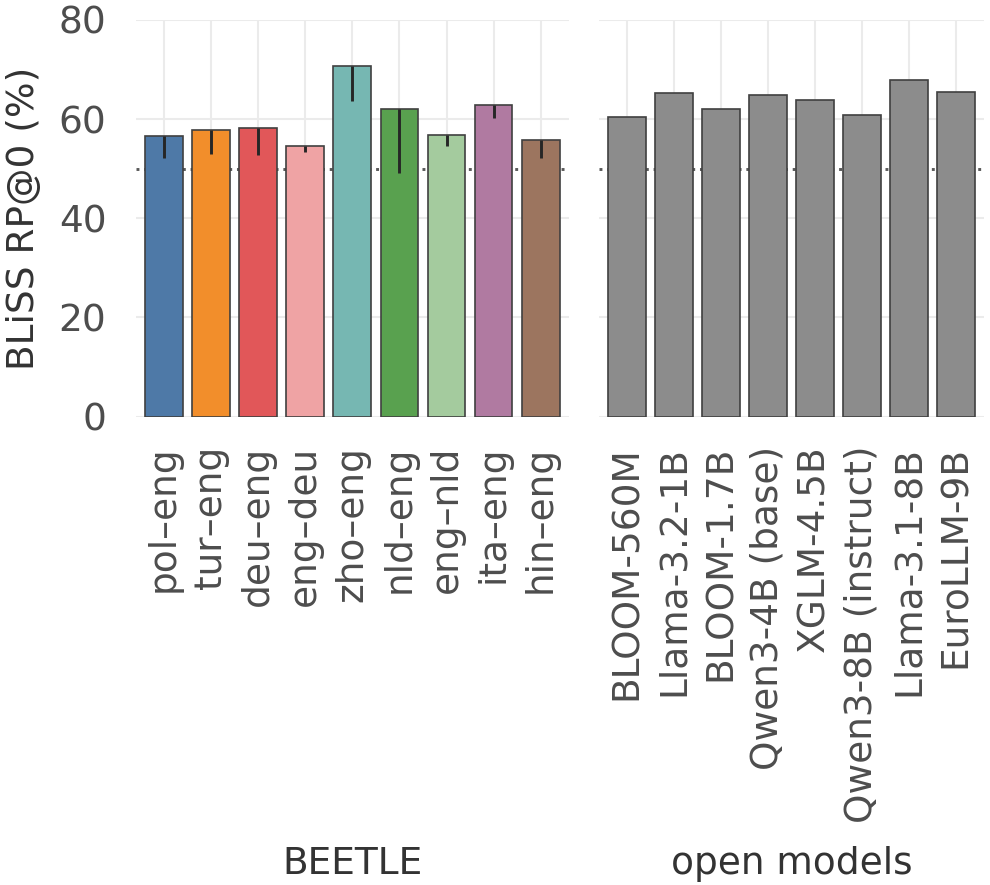}
  \caption{\textbf{BLiSS: \beetle{} vs.\ multilingual LLM performance.}
  The best \beetle{} models (zho--eng HumanScale and nld--eng FineWeb) are compared with the same baselines and a Dutch monolingual anchor (chance = 50). Large multilingual LLMs reach a higher ceiling (RP$@0 \approx 66$--$71$). Unlike on MECO, \beetle{} does not outperform the multilingual baselines on BLiSS, indicating a dissociation between the two measures. Detailed results are reported in \autoref{bliss-detailed}.}
  \label{fig:bliss_best}
\end{figure}

We secondly evaluate whether the curriculum effects observed in reading-time prediction also extend to grammatical discrimination. We use BLiSS \citep{gao2025bliss}, which tests whether models distinguish between a corrected sentence $s_{\mathrm{corr}}$, a human learner error $s_{\mathrm{lrn}}$, and an artificial error $s_{\mathrm{art}}$. We measure sentence plausibility using bits per token (BPT),

\[
\mathrm{BPT}(s) = -\frac{1}{|s|}\sum_{t=1}^{|s|} \log_2 p(w_t \mid w_{<t}).
\]

where lower BPT means that the model considers the sentence more likely. We report RP@0, the proportion of items for which the model assigns a lower BPT to the human learner error than to the corrected sentence, i.e., whether it prefers the learner error over the correction. On BLiSS, staged curricula (B2, B3-33, B5) generally outperform balanced exposure (B1), although the differences are smaller than those observed for MECO reading-time prediction (Fig.~\ref{fig:bliss_best}). On FineWeb \texttt{nld}--\texttt{eng}, the curriculum ranking is B3-33 $>$ B5 $>$ B2 $>$ B4 $>$ B1. The best curriculum also depends on the L1: B3-33 performs best for Germanic L1s, whereas B5 performs best for Chinese. Notably, BLiSS and MECO favour different curricula within \beetle{} (Spearman $\rho=-0.47$, $p<0.001$), suggesting that curriculum effects do not transfer uniformly across evaluation tasks. Overall, BLiSS shows smaller curriculum differences than MECO reading-time prediction (App. \ref{bliss-detailed}), with no single curriculum consistently dominating across languages. This suggests that the benefits of staged exposure depend on the aspect of language behaviour being evaluated. Detailed BLiSS results for the \beetle{} models and for the multilingual baselines are reported in \autoref{tab:bliss_beetle} and \autoref{tab:bliss_baselines}, and CEFR-graded pedagogical results in \autoref{tab:headline_pedagogical}.

\section{Discussion and Conclusion}\label{sec:discussion}
Although psycholinguistics and cognitive science have historically focused on monolingual English-speaking, high-resource, and WEIRD populations \citep{majid2010weird,bylund202210}, bilingualism and second-language (L2) processing are central to understanding human language processing. Computational approaches using language models inherit this bias, and an over-reliance on English and monolingual participants limits the generalisability of theories of language processing and acquisition \citep{blasi2022over}.

To address this gap, we introduced \beetle{}, a modular bilingual pretraining framework that disentangles exposure \emph{structure} from training \emph{scale} while holding architecture, tokeniser, and L2 target language fixed. Using \beetle{}, we pretrained 285 bilingual and 45 monolingual language models spanning 21 L1s and five exposure curricula, and we release checkpoints, training data, and reproducible pipelines to support future work on bilingual language modelling, learning dynamics, and cognitive modelling. \beetle{} provides interpretable and competitive models of bilingual processing. Small bilingual models trained under carefully controlled exposure conditions frequently matched or exceeded substantially larger massively multilingual systems on cognitively oriented evaluations, narrowing the gap between open-data academic models and closed-data industrial systems. Yet no single curriculum dominated. 

Sequential and late-onset schedules produced stronger alignment with human L2 reading times, suggesting that temporally structured exposure better captures incremental bilingual processing, whereas balanced exposure achieved the strongest grammaticality (BLiMP) performance, indicating that more integrated exposure may better support robust grammatical generalisation; on BLiSS error discrimination the curriculum differences were smaller, with no single schedule dominating. That different schedules produce measurably different linguistic and cognitive behaviours, even with architecture and training volume held constant, suggests that carefully structured bilingual pretraining may yield stronger computational models of human language processing and language acquisition.

By releasing a reproducible framework alongside open bilingual models designed for studying bilingualism and L2 acquisition, \beetle{} lowers the barrier to systematic research at the intersection of NLP, psycholinguistics, and cognitive science. We hope it will support future work using language models as experimentally controllable systems for investigating how language experience shapes processing, prediction, and acquisition across diverse linguistic settings.

\subsection{Future Work}
There are several features of the \beetle{} framework that we did not discuss in the paper, due to space limitations. We introduce them here and highlight their potential utility.

\paragraph{Trilingual and Multilingual Models.} \beetle{} is designed for multilingual pretraining, not just for training bilingual LMs. It can be used to develop more complex exposure curricula for multilingual LM pretraining or trilingual language modelling

\paragraph{Alternative Architectures and Additional Manipulations.} Following recent work investigating training dynamics across architectures \citep{michaelov2025language}, we implement encoder-decoder, MoE, and SSM architectures (described in App.~\ref{sec:beetle-architecture-registry}). Extending the analyses in this paper to other architectures could help us understand whether these effects are specific to transformer language models. \beetle{} additionally integrates a range of continual-learning strategies (parameter regularisation, replay, meta-learning, distillation, and architectural variants); the full registry and taxonomy are given in App.~\ref{sec:cl-strategies} (\autoref{tab:cl-registry}, \autoref{tab:taxonomy}).

\paragraph{Training Dynamics.} We trained models with rich checkpoints, but only did limited analyses of how the models changed over training across different curricula. Up to now, there have been very few models with checkpoints for any language other than English. Future work could investigate how consistent training dynamics are across languages.

\paragraph{Interpretability.}
PicoDecoder's compatibility with Pico-Analyze \citep{diehl-martinez-etal-2025-pico} allows mechanistic analysis of how bilingual pretraining manipulations affect weight geometry (e.g., shifts in effective rank or representational similarity) across checkpoints (\citealp{diehl-martinez-etal-2025-pico,inaba-etal-2025-bilingual}). 

\section{Limitations}

\paragraph{L2-English.}\label{sec:limitations-forward-direction} The \beetle~ pretraining experiments reported in the paper focus on non-English L1 data to English L2. Non-English L2s are not the focus due to limited evaluation metrics; this makes claims about directional asymmetries in transfer limited. 

\paragraph{Matched Controls and Statistical Significance.} Comparisons between \beetle{} and pretrained models, such as BLOOM-7b1, are not fully controlled for differences in parameter count, training data, tokeniser, or compute budget; a fuller control would involve training a similarly sized model (e.g., 1B parameters) on a matched FineWeb-based bilingual dataset with uniform sampling, and main results are reported using a single random seed per configuration. In particular, \beetle{} models use task-specific BPE tokenisers, whereas baselines such as BLOOM and XGLM use different tokenisation schemes, and we aggregate surprisal to the word level by summing subword contributions, which introduces a known source of noise when comparing across tokenisers \citep{pimentel-meister-2024-compute}. Within-\beetle{} comparisons are unaffected, since all compared models within a curriculum share the same tokeniser. 

\paragraph{Scope of bilingual acquisition modelling.} Our curricula attempt to provide computational analogues of several bilingual acquisition scenarios by manipulating the quantity, timing, and distribution of L1 and L2 exposure. B1 approximates relatively balanced bilingual development, with L1 and L2 available in a stable 50:50 mixture throughout training; B2--B4 approximate different forms of L2 learning, varying the onset, eventual proportion, and temporal distribution of L2 exposure; and B5 approximates late L2 learning following prolonged L1 exposure, providing a setting in which L1 attrition may occur. However, these curricula are not holistic models of bilingual acquisition: human bilingual development is shaped by factors beyond exposure history, including interaction, communicative goals, pragmatics, feedback, social context, and active language use, none of which are represented in our training setup. The curricula should therefore be understood as controlled computational analogues that isolate the role of exposure quantity and curriculum structure, rather than as comprehensive models of balanced bilingualism, L2 learning, or language attrition \citep{barbenel-etal-2026-l1}.

\paragraph{Broader Neurolinguistic and Psycholinguistic Alignment.} The current evaluation focuses on behavioural proxies such as reading-time measures and CEFR-style tasks. We do not yet evaluate alignment with neural signals such as N400 or P600 responses. Extending the analysis to neurophysiological data is left for future work.

\section{Reproducibility Statement}\label{sec:reproducibility}

All results are produced using a fully versioned and reproducible analysis pipeline. The MECO L2 evaluation follows the mixed-effects specification in Equation~\ref{eq:full-model} (with the baseline model of Equation~\ref{eq:baseline-regression}), implemented in the released analysis scripts. Models are trained on multi-node GPU infrastructure (8× A100 80GB). All checkpoints, tokenisers, and evaluation scripts are publicly released, along with the exact CSV files used to generate figures and tables. 

\section{Ethics Statement}\label{sec:ethics}

\beetle{} releases 285 pretrained bilingual and 45 monolingual L2-English language models, tokenisers, and approximately $30$ checkpoints per model. \textbf{Dataset provenance}: BabyBabelLM and FineWeb-2 are publicly released, license-compatible multilingual web corpora; we do not introduce new web crawls. We release decontaminated and pretokenized datasets for each L1-English pair across scales. The MECO~L2 eye-tracking corpus is used under its public licence with anonymous subject IDs only.  \textbf{Intended use}: \beetle{} models are scientific instruments for studying bilingual language acquisition and L2 reading dynamics. They are not deployed L2 tutors, and the CEFR-graded evaluation pipeline is for research-grade scoring rather than placement decisions. \textbf{Dual-use considerations}: the CEFR-graded text generation capability of these models could in principle be repurposed to draft L2-graded content for assessment fraud (e.g., automated CEFR-tuned essay generation). This risk is mitigated by the small parameter count of the released models (125M), which underperforms publicly-available frontier LLMs on open-ended generation; and we recommend that any downstream pedagogical-tool deployment include detection-of-machine-generated-text safeguards. \textbf{Demographic representativeness}: the released-model L1 sweep covers  typologically diverse L1s, but has limited coverage of lower-resourced L1s. 

\section{Acknowledgements}
We would like to thank CoreWeave for providing GPU resources for model training and evaluation. Suchir Salhan is supported by Cambridge University Press \& Assessment. With thanks to Laura Barbenel, Lily Goulder, Aoife O'Driscoll, Filip Trhlik, Bianca Ganescu, Diana Galván-Sosa, Gabrielle Gaudeau; Andrew Caines, and other members of the ALTA Institute and Cambridge Press \& Assessment; and Denise L\"offlad and Detmar Meurers for their support and constructive feedback on earlier stages of this work. 

\bibliography{anthology,custom}

\appendix

\IfFileExists{fig/hier/hier_macros.tex}{

\providecommand{\hierBlissCurrLoneVar}{26}
\providecommand{\hierDBfiveBoneFWoneH}{--}
\providecommand{\hierDBfiveBoneFWtfB}{30}
\providecommand{\hierDBfiveBoneFWtwoB}{-3}
\providecommand{\hierDBfiveBoneHS}{58}
\providecommand{\hierDBfiveBonePooled}{28}
\providecommand{\hierDBthreeBoneFWoneH}{--}
\providecommand{\hierDBthreeBoneFWtfB}{17}
\providecommand{\hierDBthreeBoneFWtwoB}{2}
\providecommand{\hierDBthreeBoneHS}{65}
\providecommand{\hierDBthreeBonePooled}{28}
\providecommand{\hierDBtwoBoneFWoneH}{--}
\providecommand{\hierDBtwoBoneFWtfB}{33}
\providecommand{\hierDBtwoBoneFWtwoB}{-0}
\providecommand{\hierDBtwoBoneHS}{68}
\providecommand{\hierDBtwoBonePooled}{34}
\providecommand{\hierNLone}{3}
\providecommand{\hierNrows}{59}
\providecommand{\hierPBfiveBoneFWoneH}{--}
\providecommand{\hierPBfiveBoneFWtfB}{0.87}
\providecommand{\hierPBfiveBoneFWtwoB}{0.44}
\providecommand{\hierPBfiveBoneHS}{0.99}
\providecommand{\hierPBfiveBonePooled}{0.97}
\providecommand{\hierPBthreeBoneFWoneH}{--}
\providecommand{\hierPBthreeBoneFWtfB}{0.74}
\providecommand{\hierPBthreeBoneFWtwoB}{0.53}
\providecommand{\hierPBthreeBoneHS}{1.00}
\providecommand{\hierPBthreeBonePooled}{0.97}
\providecommand{\hierPBtwoBoneFWoneH}{--}
\providecommand{\hierPBtwoBoneFWtfB}{0.90}
\providecommand{\hierPBtwoBoneFWtwoB}{0.49}
\providecommand{\hierPBtwoBoneHS}{1.00}
\providecommand{\hierPBtwoBonePooled}{0.99}
\providecommand{\hierReducedOrder}{B3-33 $>$ B1 $>$ B2}
\providecommand{\hierSigmaLone}{71.1}
\providecommand{\hierSigmaResid}{41.7}
\providecommand{\hierSigmaSeed}{7.7}
\providecommand{\hierVarCurr}{8}
\providecommand{\hierVarCurrLone}{5}
\providecommand{\hierVarLone}{29}
\providecommand{\hierVarResid}{37}
\providecommand{\hierVarScale}{21}}{}
\section*{Appendix}

\section{Model Architecture}\label{sec:beetle-architecture-registry}
 
BeetleLM is built on a custom \textsc{Pico} decoder stack \citep{diehl-martinez-etal-2025-pico} with curriculum-aware checkpointing, a registry of fifteen continual-learning strategies, and a tokeniser layer supporting per-language, shared-bilingual, and shared-trilingual vocabularies.

By default, \beetle{} uses  the \texttt{PicoDecoder}, a LLaMA-style \citep{touvron2023llama} decoder with $L=14$ blocks, RMSNorm \citep{zhang2019rms}, grouped-query attention \citep{ainslie2023gqa} (12 query heads, 1 key/value head), rotary position embeddings \citep{su2024rope}, and SwiGLU feed-forward \citep{shazeer2020glu} expanding to $4d$ and projecting back to $d$. The shared bilingual vocabulary used in the main runs has size $50{,}304$; per-language and shared-trilingual tokenisers are configured separately. Sequence length is $512$ under BF16-mixed precision. Width is the only scale-dependent hyper-parameter: $d\in\{96,384,768,1536\}$ for the tiny, small, medium, and large variants. The 125M-parameter ``medium'' variant ($d=768$, 12 heads, 1 KV head, 14 layers) is used for every reported run. 

\subsection{Alternative Architectures}

Complementary architectures enable controlled comparisons across inductive biases: \textbf{BGPT}, a GPT-2-style decoder with learned positional embeddings used in \citet{arnett-etal-2025-acquisition}; \textbf{BeetleMoE}, a top-$k$ mixture-of-experts model with ST-MoE auxiliary routing losses; and \textbf{BeetleSSM}, which replaces attention blocks with Mamba-style state-space modules. 

Pre-pretraining (Shuffle-Dyck) and time-varying ALiBi slopes can be used with different architectures. 

\subsection{Training Hyperparameters}

Training hyperparameters for reported experiments are in \autoref{tab:hyperparams}. Training was conducted on $8\times$ A100s (64GB). 100M and 2B tokens 125M models can be trained in under an hour on $1$ A100. 24B token modles take around 5 hours on $8 \times$ A100.

\begin{table}[t]
\centering
\small

\begin{tabular}{lccc}
\toprule
Hyperparameter & 100M & 2B & 24B \\
\midrule

Architecture & PicoDecoder & PicoDecoder & PicoDecoder \\
Parameters & 125M & 125M & 125M \\
Layers & 14 & 14 & 14 \\
$d_{\mathrm{model}}$ & 768 & 768 & 768 \\
Heads (Q/KV) & 12 / 1 & 12 / 1 & 12 / 1 \\
FFN size & 3072 & 3072 & 3072 \\
Sequence length & 512 & 512 & 512 \\
Vocabulary & 50K & 50K & 50K \\

\midrule

Languages & 9 & 20 & 20 \\
Dataset &
\begin{tabular}[c]{@{}c@{}}
BabyBabel \\
FineWeb-2
\end{tabular}
& FineWeb-2 & FineWeb-2 \\
\midrule

Optimizer & AdamW & AdamW & AdamW \\
Learning rate & $5\times10^{-4}$ & $5\times10^{-4}$ & $5\times10^{-4}$ \\
Weight decay & 0.01 & 0.01 & 0.01 \\
Warmup steps & 500 & 381 & 625 \\

\midrule

Micro-batch & 64 & 64 & 64 \\
Effective batch & 64 & 512 & 512 \\
Tokens / step & 32768 & 262144 & 262144 \\
Max steps & 10000 & 7629 & 91552 \\

\midrule

Precision & bf16 & bf16 & bf16 \\
Strategy & DDP & DDP & DDP \\
Devices & 8$\times$A100 & 8$\times$A100 & 8--64$\times$A100 \\

\bottomrule
\end{tabular}

\caption{Training hyperparameters across data scales.}
\label{tab:hyperparams}

\end{table}

\section{Beetle-Data}\label{sec:beetle-data}

 Beetle-Data is a module that prepares pretokenized data for language model training, streaming data from default, or user-specified, training datasets. When streaming raw documents, Beetle-Data applies a 13-gram benchmark-overlap decontaminator  \citep{brown2020language}. Reported experiments perform 13-gram decontamination for 19 evaluation metrics: BLiMP, MultiBLiMP, BLiMP-NL, ZhoBLiMP, FLORES-200, XNLI, UD Treebanks, MECO-L2 stimuli. Finally, the library pretokenises decontaminated data to user-specified training budgets and tokenization.

\section{Curriculum Parameterisation in \beetle{}}\label{app:curricula}

\beetle{} represents bilingual and multilingual curricula as continuous
exposure schedules rather than as discrete training phases with fixed
boundaries. Formally, each curriculum defines a time-varying language mixture
\(
w(t) \in \Delta^{|L|}
\)
over normalised training progress
\(
t \in [0,1],
\)
where \(\Delta^{|L|}\) denotes the simplex over the set of languages \(L\).
Progress \(t\) may be measured either in optimisation steps or in tokens seen.
At every point during training, \(w_l(t)\) specifies the probability mass
assigned to language \(l\), with
\(
\sum_l w_l(t)=1.
\)

All curriculum conditions are constructed compositionally from four schedule primitives, summarised in \autoref{tab:taxonomy}. 

\paragraph{SigmoidRamp.}
The basic transition mechanism is a smooth sigmoid interpolation between two
language-mixture vectors. The transition is parameterised by a midpoint
\(m \in [0,1]\), which determines the onset timing, and a steepness parameter
\(k\), which controls transition sharpness. Larger values of \(k\) approximate
an abrupt step function, while smaller values yield gradual transitions. Unless
otherwise stated, experiments use \(k=12\). All sigmoid-based curricula
(including B2, B3, and B5) share the same implementation, meaning that
transition sharpness is controlled by a single scalar parameter rather than by
curriculum-specific logic.

\paragraph{MultiSigmoid.}
More complex staged introductions are implemented by chaining multiple sigmoid
ramps. This permits schedules such as
\(
L_1 \rightarrow L_2 \rightarrow L_3
\)
with independently parameterised transition points. The trilingual curricula
(T2 and T3) use this mechanism.

\paragraph{BurstSchedule.}
Burst schedules superimpose periodic square-wave increases in exposure for a
target language on top of an existing base schedule. Each burst process is
parameterised by an amplitude (the additive increase in exposure weight), a
period (cycle length as a fraction of total training), a duty cycle (the
fraction of each cycle during which the burst is active), and a start time
\(t_{\mathrm{start}}\). During an active burst, the target language receives an
increased allocation and the remaining language weights are proportionally
renormalised so that
\(
\sum_l w_l(t)=1.
\)
Burst schedules are intended to model concentrated episodes of exposure, such
as intensive classroom interaction or temporary re-exposure to the first
language. They are used in conditions such as B2\_burst and the L1
re-exposure manipulation in B6.

\paragraph{CyclicalSchedule.}
Where smoother recurring fluctuations are theoretically motivated, \beetlelm{}
uses sinusoidal oscillations rather than discrete square-wave bursts. The
cyclical schedule is parameterised by an amplitude and frequency and produces
continuous oscillatory variation around a base exposure schedule.

\subsection{Episodic Clustering}
\label{app:episodic}

In addition to continuous exposure scheduling, \beetlelm{} implements episodic
clustering through a separate mechanism termed the \texttt{EpisodicSampler}.
This mechanism is orthogonal to the global exposure schedule \(w(t)\). Whereas
bursts modify the overall language proportions through time, episodic
clustering changes the local temporal structure of sampling.

The episodic sampler operates over a finite set of named contexts
\(
C = \{c_1,\ldots,c_n\},
\)
each associated with a characteristic language distribution and a marginal
sampling probability \(\pi_c\). Example contexts include
\texttt{home}
with distribution
\(
\{L_1:0.95, L_2:0.05\}
\)
and
\texttt{classroom}
with distribution
\(
\{L_1:0.3, L_2:0.7\}.
\)

Training examples are sampled in contiguous blocks of size \(B\). Within a
block, all examples are drawn from the same context distribution. This breaks
the IID assumption typically used in standard language-model training and
approximates the contextual non-stationarity characteristic of naturalistic
second-language exposure.

When both a global schedule and episodic sampling are active, the effective
per-step language weights are computed as the geometric mean of the scheduled
distribution \(w(t)\) and the currently active context distribution, followed
by renormalisation.

Condition B4 (\emph{Classroom}) combines both mechanisms: an underlying
sigmoid transition toward an 80:20 language mixture, periodic L2 bursts, and a
two-context episodic sampler consisting of \texttt{home} (80\%) and
\texttt{classroom} (20\%) contexts with block size \(B=64\). Condition B1b is
a corresponding non-classroom control using a balanced 50:50 base schedule
with episodic clustering only. B4a and B4b are ablations removing episodic
clustering and EWC respectively.

\begin{table*}[!ht]
\centering
\small
\setlength{\tabcolsep}{3.5pt}
\resizebox{\textwidth}{!}{%
\begin{tabular}{llll}
\toprule
\textbf{Family} & \textbf{Strategy} & \textbf{Mechanism} & \textbf{Key hyperparameters} \\
\midrule
\multicolumn{4}{l}{\textit{Parameter regularisation}} \\
& EWC \citep{kirkpatrick2017overcoming} & Diagonal-Fisher quadratic penalty at phase change & $\lambda\in[0,40]$, $K_{\text{Fisher}}=10$ \\
& SI & Online importance accumulation & $c$, damping \\
& MAS & Gradient-magnitude importance & $\lambda$ \\
\midrule
\multicolumn{4}{l}{\textit{Replay}} \\
& LAMOL \citep{sun2019lamol} & Pseudo-sample generation, nucleus sampling & $\lambda_{\text{replay}}\in[0,0.5]$, top-$p$ \\
& InsCL similarity replay & Centroid-similarity weighted replay & cosine/Wasserstein, buffer size \\
& Episodic replay & Stored exemplars, uniform sampling & buffer size, mix ratio \\
\midrule
\multicolumn{4}{l}{\textit{Meta-learning}} \\
& MAML / SMLMT & Inner-loop adaptation on $(S,Q)$ splits & hybrid ratio $\in[0,0.4]$, $N$-way, $k$-shot \\
\midrule
\multicolumn{4}{l}{\textit{Distillation}} \\
& Hinton-KD & Soft-target cross-entropy against pre-L2 anchor & $T$, $\lambda_{KD}$ \\
& Self-distillation & Same model as teacher at earlier checkpoint & checkpoint offset \\
& FitNets feature distill. & Hidden-state $\ell_2$ alignment & layer-pair list \\
\midrule
\multicolumn{4}{l}{\textit{Architectural / auxiliary loss}} \\
& TILT freeze-and-extend & Freeze backbone; train embeddings + LM head & frozen modules \\
& Exposure-LR plasticity & Token-exposure LR with phase-transition boost & boost $\in[1,3]$, warmup tokens \\
& ALiBi scheduler & Time-varying ALiBi slopes across phases & initial/final slope, decay shape \\
& Shuffle-Dyck pre-pretraining & Formal-language warmup before NL & steps, depth \\
& EWC + MAML stack & Composite of regularisation + meta-learning & paired hyperparameters \\
\bottomrule
\end{tabular}%
}
\caption{The continual-learning strategies implemented in the \beetle{} training library. The experiments reported here use EWC, LAMOL, and exposure-LR plasticity overlays.}
\label{tab:cl-registry}
\end{table*}

A bilingual curriculum is a time-dependent mixture $\mathcal{D}_{\text{mix}}(t) = \alpha(t)\,\mathcal{D}_1 \cup (1-\alpha(t))\,\mathcal{D}_2$, where $\alpha(t)\in[0,1]$ is the per-step L2 sampling probability and $t\in[0,T]$ indexes training. The polyglot generalisation is a language-exposure vector $\boldsymbol{\alpha}(t)=(\alpha_1(t),\ldots,\alpha_N(t))$ with $\sum_i \alpha_i(t)=1$. Each schedule maps a normalised progress value $p\in[0,1]$ to a per-language sampling weight; phase boundaries are discrete events that trigger EWC snapshots in continual-learning runs. Rather than specifying $\boldsymbol{\alpha}(t)$ as an arbitrary continuous function, we represent curricula as a sequence of discrete phases $\mathcal{C} = \{(\tau_k^{\text{start}}, \tau_k^{\text{end}}, \mathbf{w}_k)\}_{k=1}^{K}$, where $\mathbf{w}_k \in \mathbb{R}^N$ satisfies $\sum_i w_{k,i}=1$; for $\tau_k^{\text{start}} \le t/T < \tau_k^{\text{end}}$ we set $\boldsymbol{\alpha}(t) = \mathbf{w}_k$. This declarative phase representation is used in every reported run.
 
\begin{table}[t]
\centering
\small
\setlength{\tabcolsep}{3pt}
\begin{tabular}{lccccc}
\toprule
& \textbf{B1} & \textbf{B2} & \textbf{B3} & \textbf{B4} & \textbf{B5} \\
\midrule
L2 onset      & 0   & 50\% & 50\% & 0   & 80\% \\
Final L2      & 50\% & 50\% & 67\% & 20\% & 50\% \\
Transition    & none & sig. & sig. & bursts & sig. \\
Bursting      & no   & no   & no   & yes  & no \\
Episodic      & no   & no   & no   & yes  & no \\
CL overlay    & none & none & none & none & none \\
\bottomrule
\end{tabular}
\caption{Controlled curriculum factors for bilingual conditions B1--B5. B4 uses bursty episodic exposure with the same mean L2 ratio as B2 and no EWC.}
\label{tab:taxonomy}
\end{table}

\subsection{\beetle{} Continual-Learning Strategies}\label{sec:cl-strategies}
 
The BeetleLM library contains  \textbf{fifteen} modular continual-learning strategies, which are grouped in five families, summarised in Tab.~\ref{tab:cl-registry}.

\emph{Parameter regularisation}: EWC \citep{kirkpatrick2017overcoming} with $\lambda\in[0,40]$ and a Fisher snapshot at phase change, SI (synaptic intelligence), and MAS (memory-aware synapses). 

\emph{Generative and similarity replay}: LAMOL \citep{sun2019lamol} with $\lambda_{\text{replay}}\in[0,0.5]$, InsCL-style similarity replay (cosine and Wasserstein centroids), and episodic-sample replay buffers. 

\emph{Meta-learning}: first-order MAML / SMLMT \citep{bansal-etal-2020-self,africa-etal-2025-learning,africa-etal-2025-meta} with hybrid ratio $\in[0,0.4]$ and $N$-way/$k$-shot inner episodes. 

\emph{Distillation}: feature-level (FitNets), output-level (Hinton-KD), and self-distillation against the pre-L2 anchor. \emph{Architectural and auxiliary-loss}: TILT freeze-and-extend, exposure-LR plasticity boost (\texttt{plasticity\_boost}$\in[1,3]$), ALiBi-scheduler decay, and Shuffle-Dyck pre-pretraining.

\section{Bilingual and Second Language Evaluation Methodology}\label{app:pmeco-future}

We select different three evaluation families probe distinct facets of second-language competence. MECO measures graded processing alignment through reading-time prediction. BLiSS \citep{gao2025bliss} tests categorical grammatical knowledge, and its triplet design---contrasting a corrected form, a genuine learner error, and a synthetic error---additionally asks whether a model internalises an L2-specific interlanguage error structure rather than a generic notion of ungrammaticality. JFLEG \citep{napoles-etal-2017-jfleg} tests whether a model prefers corrected forms over learner productions, connecting the framework to grammatical error correction. Because \beetle{} employs a shared vocabulary under controlled exposure schedules, it offers a platform for studying bilingual-specific phenomena that lie beyond monolingual processing. Candidate directions include lexical facilitation and cognate effects of the kind modelled by \citet{dijkstra2002architecture}, the resolution of interlingual homographs, code-switching and the switch costs documented in the bilingual production and comprehension literature \citep{meuter1999bilingual,gollan2009should}, and cross-linguistic syntactic interference \citep{hartsuiker2004crosslinguistic}. These phenomena are noted here as future directions that the controlled-exposure design of the framework is well placed to support.

\section{Details of \beetle{} MECO Analyses}\label{sec:meco-results}

For each target word $w_i$, we compute autoregressive surprisal
$S(w_i) = -\log P(w_i \mid w_{<i})$ on the relevant \beetle{} and baseline models. Following previous work,
 \citep[e.g.,][]{wilcox-etal-2023-language}, we fit mixed-effects regressions
predicting first-pass reading time from current-word and spillover
surprisal while controlling for word length,
frequency, and sentence position, and evaluate language model alignment with reading time based on the fit of the regressions to the data. 

We fit a linear mixed-effects regression models to MECO
first-pass duration (also known as \textit{gaze duration}) data from MECO waves 1 + 2 based on our predictors. We first fit baseline regressions with each word's position in the sentence, length in characters, and frequency as main effects, as well as each the preceding two words' length and frequency; and include by-subject uncorrelated random slopes for each of these as well as random intercepts for each subject:
\begin{align}
\label{eq:baseline-regression}
\mathrm{first\_pass} \sim{} &\mathrm{position}(w_i)+ \mathrm{len}(w_i)\nonumber\\&+ \mathrm{len}(w_{i-1}) + \mathrm{len}(w_{i-2})\nonumber\\
&+ \mathrm{freq}(w_i) + \mathrm{freq}(w_{i-1})\nonumber\\
& +\mathrm{freq}(w_{i-2})\nonumber\\
&+ (1 +\mathrm{position}(w_i)+ \mathrm{len}(w_i)\nonumber\\
&~~~~~~~~+ \mathrm{len}(w_{i-1}) + \mathrm{len}(w_{i-2})\nonumber\\
&~~~~~~~~+ \mathrm{freq}(w_i) + \mathrm{freq}(w_{i-1})\nonumber\\
&~~~~~~~~+ \mathrm{freq}(w_{i-2})\nonumber \mid\mid \mathrm{subject}) \\
\end{align}

We then fit linear mixed-effects regressions that included all the baseline predictors in addition to the surprisal of the current and previous two words, which were also included as uncorrelated random slopes:

\begin{align}
\label{eq:full-model}
\mathrm{first\_pass} \sim{} &\mathrm{Surprisal}(w_i)\nonumber\\
&+ \mathrm{Surprisal}(w_{i-1})\nonumber\\&+ \mathrm{Surprisal}(w_{i-w}) \nonumber\\
&+ (1 +\mathrm{Surprisal}(w_i)\nonumber\\
&~~~~~~~~+ \mathrm{Surprisal}(w_{i-1})\nonumber\\&~~~~~~~~+\mathrm{Surprisal}(w_{i-w}) \mid\mid \mathrm{subject})\nonumber\\
&+ \mathrm{baseline\_predictors} \nonumber\\
\end{align}

Since we hold all regression predictors constant except the surprisal terms, we consider language models with surprisal terms that lead to better fits (i.e., higher log-likelihoods) to first-pass duration to be models that better align with reading time. For easier comparison, we subtract the log-likelihood of the baseline regressions to get $\Delta\log L$: 

\begin{equation}
\Delta\log L = \log L(\text{full}) - \log L(\text{baseline}).
\end{equation}

We thus compare language model alignment with reading time based on the $\Delta\log L$ of regressions that include surprisal calculated using them.

The script used to fit the regressions is released with the code. Sample
sizes are on the order of tens of subjects and tens of thousands of
word-level observations.

\section{BLiSS and Pedagogical Evaluation}

We evaluate whether curriculum effects observed in reading-time prediction extend to grammatical discrimination and pedagogical benchmarks. First, we consider BLiSS \citep{gao2025bliss}, which evaluates preference behaviour over L2 error triplets, consisting of a corrected sentence $s_{\mathrm{corr}}$, a human learner error $s_{\mathrm{lrn}}$, and an artificial error $s_{\mathrm{art}}$. We compute token-normalized surprisal:
\[
\mathrm{BPT}(s) = -\frac{1}{|s|}\sum_{t=1}^{|s|}\log_2 p(w_t \mid w_{<t}),
\]
where lower values indicate higher model plausibility.

We report four metrics following \citep{gao2025bliss}. \textbf{Learner Preference (LP)} measures whether $s_{\mathrm{lrn}}$ is preferred over $s_{\mathrm{corr}}$ ($\mathrm{BPT}(s_{\mathrm{lrn}}) < \mathrm{BPT}(s_{\mathrm{corr}})$). This metric is informative but does not distinguish between grammatical competence and learner simulation behavior. \textbf{Human vs. Artificial Preference (HAP)} measures whether $s_{\mathrm{lrn}}$ is preferred over $s_{\mathrm{art}}$, and \textbf{HAP-$\tau$} applies a margin constraint for robustness. \textbf{Strict Order (SO)} requires the full ordering $s_{\mathrm{corr}} \prec s_{\mathrm{lrn}} \prec s_{\mathrm{art}}$. We report all metrics separately as they capture different aspects of model behavior.

\subsection{Detailed \beetle{} BLiSS Results}\label{bliss-detailed}

\begin{figure*}[!ht]
    \centering
    \includegraphics[width=1\linewidth]{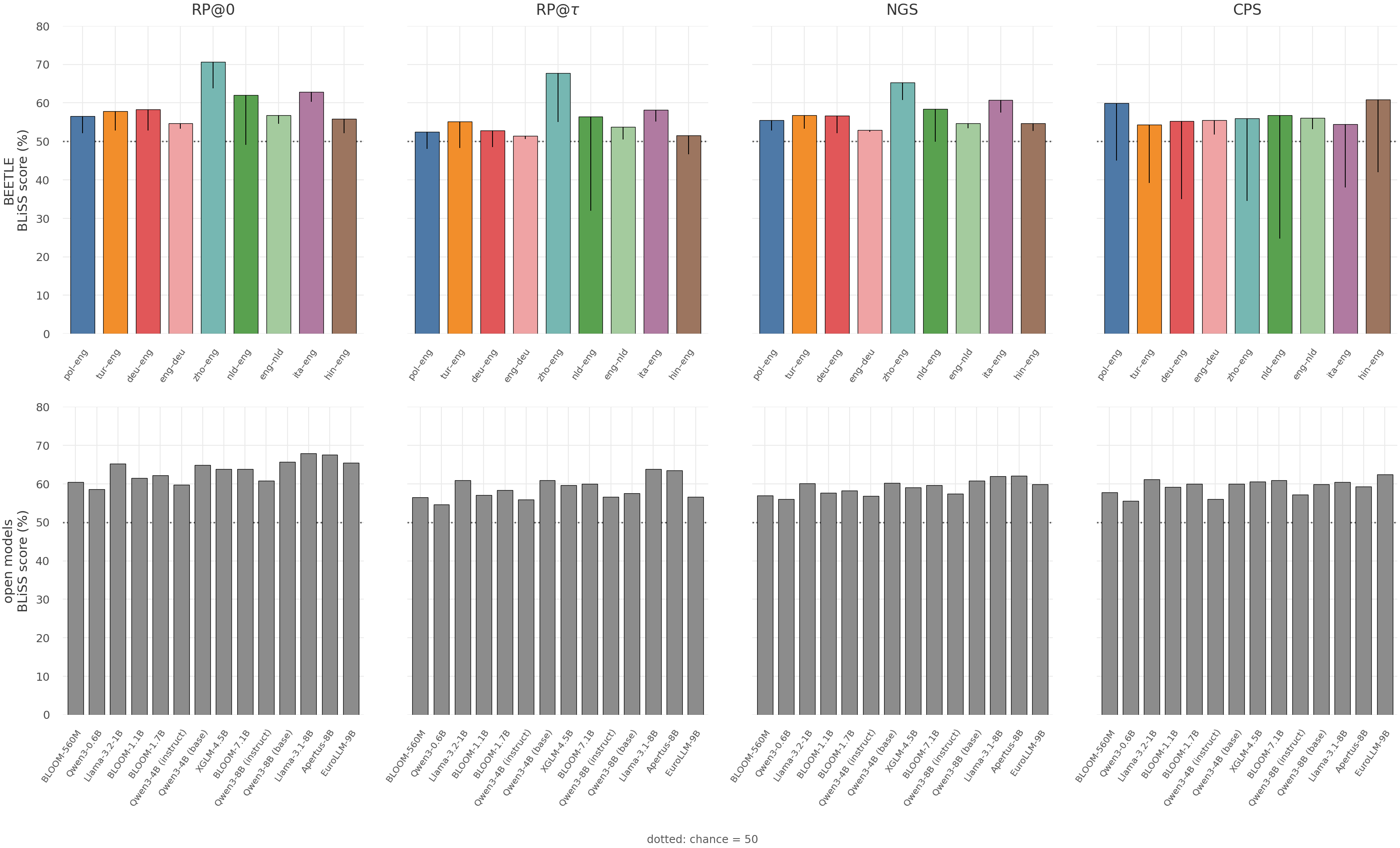}
    \caption{Overall BLiSS learner-minimal-pair accuracy for \beetle{} models against baseline-matched multilingual LLMs, aggregated across the matched-L1 cohorts. Curriculum differences on BLiSS are smaller than the corresponding differences in MECO reading-time alignment.}
    \label{fig:bliss-apdx-main}
\end{figure*}

\autoref{tab:bliss_beetle} and \autoref{tab:bliss_baselines} show BLiSS accuracy of \beetle{} and LLMs. The overall performance is summarised in \textit{Figure} \ref{fig:bliss-apdx-main}. Beyond this headline accuracy, the detailed BLiSS table (\autoref{tab:bliss_beetle}) decomposes performance into RP@0 and RP@$\tau$ (random-over-human preference, with and without a margin), NGS (normalised gap score), and CPS (correct-preferred sanity check); reporting these additional metrics separates whether models merely prefer human over random productions (RP@0) from whether they do so by a robust margin (RP@$\tau$), and confirms the effect is not a scoring artefact (CPS/NGS), giving a finer-grained picture than the headline accuracy in \autoref{fig:bliss-apdx-main}. We briefly summarise the main results.

\paragraph{Curriculum effects on BLiSS}On BLiSS, staged curricula (B2, B3-33, B5) outperform B1, but effect sizes are smaller than those observed on MECO (Fig.~\ref{fig:bliss_best}). On FineWeb \texttt{nld}--\texttt{eng}, the ordering is B3-33 $>$ B5 $>$ B2 $>$ B4 $>$ B1 , compared to a larger separation in MECO (approximately 78 $\Delta \log L$). The best-performing curriculum varies by L1: B3-33 ranks highest for Germanic L1s, while B5 ranks highest for Chinese L1. Within \beetle{}, BLiSS and MECO induce different curriculum rankings (Spearman $\rho=-0.47$, $p<0.001$).

\paragraph{Summary across discrimination tasks} Across BLiSS, curriculum differences are smaller than those observed for MECO reading-time prediction (\autoref{bliss-detailed}).  Best-performing curricula vary by L1 (B3-33 for Germanic, B5 for Chinese), with no single schedule dominating across settings.  Across pedagogical evaluations (\autoref{tab:headline_pedagogical}), balanced exposure (B1) consistently performs best on cloze and CEFR-style metrics. In contrast, staged curricula only dominate on MECO reading-time alignment, indicating a clear separation between fluency/correction-based judgments and online processing measures.

\subsection{CEFR Task Performance}

This section reports \beetle{}'s performance on the CEFR-aligned and other pedagogical evaluations.

\begin{figure*}[!ht]
    \centering
    \includegraphics[width=1\linewidth]{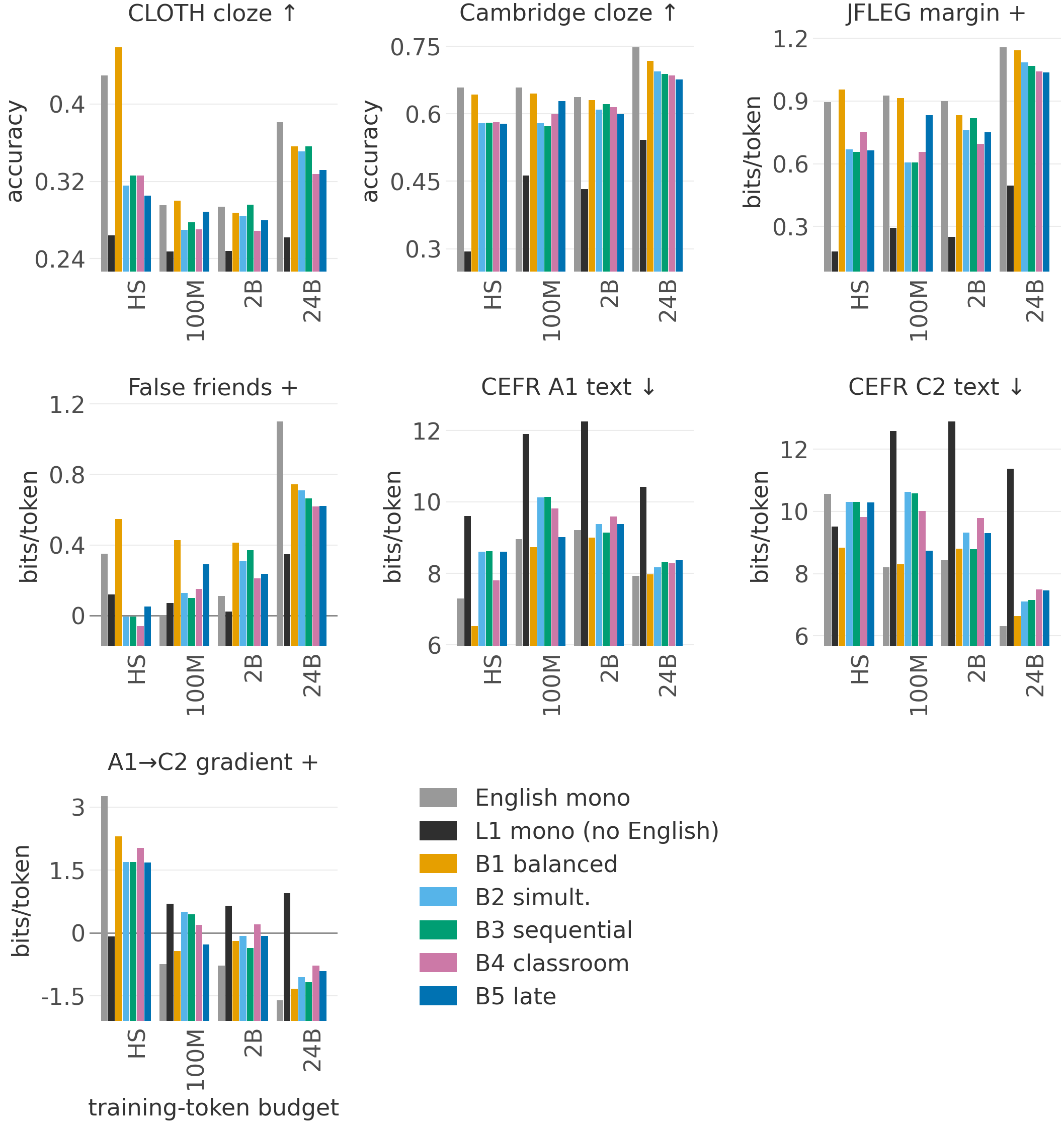}
    \caption{Performance of \beetle{} 125M models across pedagogical evaluations (CLOTH and Cambridge cloze, JFLEG grammatical correction, false-friend disambiguation, and the CEFR perplexity gradient), by curriculum and training scale. Balanced exposure (B1) is consistently strongest on these fluency- and correction-based measures.}
    \label{fig:pedagogical}
\end{figure*}

Fig. \ref{fig:pedagogical} shows performance of \beetle{} on different pedagogical tasks. Tab.~\ref{tab:headline_pedagogical} summarises the downstream performance of \beetle{} 125M models on CLOTH closed-cloze accuracy, Cambridge open-cloze accuracy, JFLEG learner-fluency margin, false-friend disambiguation margin, and the CEFR perplexity gradient, by curriculum and training scale. The individual benchmarks measure the following:
\begin{itemize}
  \item \textbf{CLOTH} \citep{xie-etal-2018-cloth}: a large cloze test of English built from middle/high-school exam passages; closed multiple-choice gap-filling measuring grammatical and lexical competence.
  \item \textbf{Cambridge cloze}: open-cloze items from Cambridge English exams graded by CEFR level; the model must supply the missing word, testing productive grammatical knowledge \citep{mullooly2023cupa5}.
  \item \textbf{JFLEG grammatical correction} \citep{napoles-etal-2017-jfleg}: a fluency/grammatical-error-correction benchmark; we score the model's preference (bits/token margin) for the corrected sentence over the learner-error version (positive = prefers the correction).
  \item \textbf{False-friend disambiguation}: cognate/false-friend minimal pairs; the margin (bits/token) by which the model prefers the contextually correct sense over the interlingual-homograph distractor.
  \item \textbf{CEFR perplexity gradient} \citep{arase-etal-2022-cefr}: the monotonic change in bits-per-token from CEFR-A1 to CEFR-C2 graded texts; a well-calibrated model should assign lower perplexity to easier (A1) than harder (C2) material, so the A1$\to$C2 gradient indexes proficiency sensitivity.
\end{itemize}

\clearpage 

\begin{table*}[!ht]
\centering\scriptsize
\setlength{\tabcolsep}{3pt}
\begin{minipage}[t]{0.49\linewidth}\centering
\begin{tabular}{ll l cccc}
\toprule
& & & \multicolumn{4}{c}{\textbf{BLiSS (matched-L1 cohort)}} \\
\cmidrule(lr){4-7}
\textbf{Pair} & \textbf{Data} & \textbf{Curr.} & \textbf{RP@0} & \textbf{RP@$\tau$} & \textbf{NGS} & \textbf{CPS} \\
\midrule
\texttt{deu--eng} & Human-scale & Mono & \cellcolor[rgb]{0.990,1.000,0.990} 51.0 & \cellcolor[rgb]{1.000,0.887,0.887} 38.3 & \cellcolor[rgb]{0.983,1.000,0.983} 51.3 & \cellcolor[rgb]{1.000,0.774,0.774} 29.5 \\
 & Human-scale & B1 balanced & \cellcolor[rgb]{0.925,1.000,0.925} \textbf{57.2} & \cellcolor[rgb]{1.000,0.983,0.983} \textbf{48.2} & \cellcolor[rgb]{0.927,1.000,0.927} 55.5 & \cellcolor[rgb]{1.000,0.965,0.965} 46.8 \\
 & Human-scale & B2 simult. & \cellcolor[rgb]{0.950,1.000,0.950} 54.8 & \cellcolor[rgb]{1.000,0.947,0.947} 44.5 & \cellcolor[rgb]{0.935,1.000,0.935} 54.9 & \cellcolor[rgb]{1.000,0.836,0.836} 35.1 \\
 & Human-scale & B3 sequential & \cellcolor[rgb]{0.943,1.000,0.943} 55.5 & \cellcolor[rgb]{1.000,0.971,0.971} 47.0 & \cellcolor[rgb]{0.927,1.000,0.927} 55.5 & \cellcolor[rgb]{1.000,0.853,0.853} 36.7 \\
 & Human-scale & B4 classroom & \cellcolor[rgb]{0.950,1.000,0.950} 54.8 & \cellcolor[rgb]{1.000,0.942,0.942} 44.0 & \cellcolor[rgb]{0.938,1.000,0.938} 54.7 & \cellcolor[rgb]{1.000,0.882,0.882} 39.3 \\
 & Human-scale & B5 late & \cellcolor[rgb]{0.946,1.000,0.946} 55.2 & \cellcolor[rgb]{1.000,0.951,0.951} 44.9 & \cellcolor[rgb]{0.939,1.000,0.939} 54.6 & \cellcolor[rgb]{1.000,0.839,0.839} 35.4 \\
\addlinespace[1.5pt]
 & 100M tok & B1 balanced & \cellcolor[rgb]{0.952,1.000,0.952} 54.6 & \cellcolor[rgb]{1.000,0.960,0.960} 45.8 & \cellcolor[rgb]{0.942,1.000,0.942} 54.4 & \cellcolor[rgb]{0.998,1.000,0.998} 50.1 \\
 & 100M tok & B2 simult. & \cellcolor[rgb]{0.942,1.000,0.942} 55.6 & \cellcolor[rgb]{1.000,0.972,0.972} 47.1 & \cellcolor[rgb]{0.918,1.000,0.918} \textbf{56.2} & \cellcolor[rgb]{1.000,0.891,0.891} 40.1 \\
 & 100M tok & B3 sequential & \cellcolor[rgb]{0.942,1.000,0.942} 55.6 & \cellcolor[rgb]{1.000,0.955,0.955} 45.3 & \cellcolor[rgb]{0.934,1.000,0.934} 55.0 & \cellcolor[rgb]{1.000,0.870,0.870} 38.2 \\
 & 100M tok & B4 classroom & \cellcolor[rgb]{0.949,1.000,0.949} 54.9 & \cellcolor[rgb]{1.000,0.957,0.957} 45.5 & \cellcolor[rgb]{0.925,1.000,0.925} 55.7 & \cellcolor[rgb]{1.000,0.934,0.934} 44.0 \\
 & 100M tok & B5 late & \cellcolor[rgb]{0.936,1.000,0.936} 56.1 & \cellcolor[rgb]{1.000,0.972,0.972} 47.1 & \cellcolor[rgb]{0.930,1.000,0.930} 55.3 & \cellcolor[rgb]{1.000,0.979,0.979} 48.1 \\
\addlinespace[1.5pt]
 & 2B tok & B1 balanced & \cellcolor[rgb]{0.964,1.000,0.964} 53.5 & \cellcolor[rgb]{1.000,0.937,0.937} 43.5 & \cellcolor[rgb]{0.955,1.000,0.955} 53.4 & \cellcolor[rgb]{1.000,0.986,0.986} 48.7 \\
 & 2B tok & B2 simult. & \cellcolor[rgb]{0.943,1.000,0.943} 55.5 & \cellcolor[rgb]{1.000,0.962,0.962} 46.1 & \cellcolor[rgb]{0.934,1.000,0.934} 55.0 & \cellcolor[rgb]{1.000,0.967,0.967} 47.0 \\
 & 2B tok & B3 sequential & \cellcolor[rgb]{0.964,1.000,0.964} 53.5 & \cellcolor[rgb]{1.000,0.960,0.960} 45.8 & \cellcolor[rgb]{0.951,1.000,0.951} 53.7 & \cellcolor[rgb]{1.000,0.999,0.999} 49.9 \\
 & 2B tok & B4 classroom & \cellcolor[rgb]{0.941,1.000,0.941} 55.7 & \cellcolor[rgb]{1.000,0.954,0.954} 45.2 & \cellcolor[rgb]{0.926,1.000,0.926} 55.6 & \cellcolor[rgb]{1.000,0.934,0.934} 44.0 \\
 & 2B tok & B5 late & \cellcolor[rgb]{0.937,1.000,0.937} 56.0 & \cellcolor[rgb]{1.000,0.949,0.949} 44.7 & \cellcolor[rgb]{0.931,1.000,0.931} 55.2 & \cellcolor[rgb]{1.000,0.978,0.978} 48.0 \\
\addlinespace[1.5pt]
 & 24B tok & B1 balanced & \cellcolor[rgb]{0.968,1.000,0.968} 53.1 & \cellcolor[rgb]{1.000,0.969,0.969} 46.8 & \cellcolor[rgb]{0.960,1.000,0.960} 53.0 & \cellcolor[rgb]{0.959,1.000,0.959} \textbf{52.0} \\
 & 24B tok & B2 simult. & \cellcolor[rgb]{0.949,1.000,0.949} 54.9 & \cellcolor[rgb]{1.000,0.964,0.964} 46.3 & \cellcolor[rgb]{0.926,1.000,0.926} 55.6 & \cellcolor[rgb]{1.000,0.957,0.957} 46.1 \\
\midrule
\texttt{nld--eng} & Human-scale & Mono & \cellcolor[rgb]{0.945,1.000,0.945} 55.3 & \cellcolor[rgb]{1.000,0.948,0.948} 44.6 & \cellcolor[rgb]{0.939,1.000,0.939} 54.6 & \cellcolor[rgb]{1.000,0.805,0.805} 32.3 \\
 & Human-scale & B1 balanced & \cellcolor[rgb]{0.931,1.000,0.931} 56.6 & \cellcolor[rgb]{1.000,0.989,0.989} 48.9 & \cellcolor[rgb]{0.931,1.000,0.931} 55.2 & \cellcolor[rgb]{1.000,0.982,0.982} 48.4 \\
 & Human-scale & B2 simult. & \cellcolor[rgb]{0.922,1.000,0.922} 57.5 & \cellcolor[rgb]{0.999,1.000,0.999} 50.1 & \cellcolor[rgb]{0.902,1.000,0.902} 57.4 & \cellcolor[rgb]{1.000,0.874,0.874} 38.6 \\
 & Human-scale & B3 sequential & \cellcolor[rgb]{0.920,1.000,0.920} 57.7 & \cellcolor[rgb]{1.000,0.991,0.991} 49.1 & \cellcolor[rgb]{0.907,1.000,0.907} 57.0 & \cellcolor[rgb]{1.000,0.893,0.893} 40.3 \\
 & Human-scale & B4 classroom & \cellcolor[rgb]{0.924,1.000,0.924} 57.3 & \cellcolor[rgb]{1.000,0.977,0.977} 47.6 & \cellcolor[rgb]{0.923,1.000,0.923} 55.8 & \cellcolor[rgb]{1.000,0.915,0.915} 42.3 \\
 & Human-scale & B5 late & \cellcolor[rgb]{0.926,1.000,0.926} 57.1 & \cellcolor[rgb]{1.000,0.987,0.987} 48.6 & \cellcolor[rgb]{0.911,1.000,0.911} 56.7 & \cellcolor[rgb]{1.000,0.881,0.881} 39.2 \\
\addlinespace[1.5pt]
 & 100M tok & B1 balanced & \cellcolor[rgb]{0.926,1.000,0.926} 57.1 & \cellcolor[rgb]{0.999,1.000,0.999} 50.1 & \cellcolor[rgb]{0.905,1.000,0.905} 57.2 & \cellcolor[rgb]{1.000,0.988,0.988} 48.9 \\
 & 100M tok & B2 simult. & \cellcolor[rgb]{0.901,1.000,0.901} 59.5 & \cellcolor[rgb]{0.997,1.000,0.997} 50.2 & \cellcolor[rgb]{0.890,1.000,0.890} \textbf{58.3} & \cellcolor[rgb]{1.000,0.898,0.898} 40.7 \\
 & 100M tok & B3 sequential & \cellcolor[rgb]{0.903,1.000,0.903} 59.3 & \cellcolor[rgb]{0.995,1.000,0.995} 50.4 & \cellcolor[rgb]{0.903,1.000,0.903} 57.3 & \cellcolor[rgb]{1.000,0.893,0.893} 40.3 \\
 & 100M tok & B4 classroom & \cellcolor[rgb]{0.918,1.000,0.918} 57.9 & \cellcolor[rgb]{1.000,0.998,0.998} 49.8 & \cellcolor[rgb]{0.905,1.000,0.905} 57.2 & \cellcolor[rgb]{1.000,0.906,0.906} 41.5 \\
 & 100M tok & B5 late & \cellcolor[rgb]{0.926,1.000,0.926} 57.1 & \cellcolor[rgb]{1.000,0.990,0.990} 49.0 & \cellcolor[rgb]{0.909,1.000,0.909} 56.9 & \cellcolor[rgb]{1.000,0.987,0.987} 48.8 \\
\addlinespace[1.5pt]
 & 2B tok & B1 balanced & \cellcolor[rgb]{0.911,1.000,0.911} 58.5 & \cellcolor[rgb]{0.992,1.000,0.992} 50.6 & \cellcolor[rgb]{0.907,1.000,0.907} 57.0 & \cellcolor[rgb]{1.000,0.975,0.975} 47.7 \\
 & 2B tok & B2 simult. & \cellcolor[rgb]{0.923,1.000,0.923} 57.4 & \cellcolor[rgb]{0.997,1.000,0.997} 50.2 & \cellcolor[rgb]{0.898,1.000,0.898} 57.7 & \cellcolor[rgb]{1.000,0.885,0.885} 39.6 \\
 & 2B tok & B3 sequential & \cellcolor[rgb]{0.916,1.000,0.916} 58.1 & \cellcolor[rgb]{1.000,0.991,0.991} 49.1 & \cellcolor[rgb]{0.897,1.000,0.897} 57.8 & \cellcolor[rgb]{1.000,0.860,0.860} 37.3 \\
 & 2B tok & B4 classroom & \cellcolor[rgb]{0.907,1.000,0.907} 58.9 & \cellcolor[rgb]{1.000,0.999,0.999} 49.9 & \cellcolor[rgb]{0.898,1.000,0.898} 57.7 & \cellcolor[rgb]{1.000,0.881,0.881} 39.2 \\
 & 2B tok & B5 late & \cellcolor[rgb]{0.909,1.000,0.909} 58.7 & \cellcolor[rgb]{0.995,1.000,0.995} 50.4 & \cellcolor[rgb]{0.899,1.000,0.899} 57.6 & \cellcolor[rgb]{1.000,0.851,0.851} 36.5 \\
\addlinespace[1.5pt]
 & 24B tok & B1 balanced & \cellcolor[rgb]{0.931,1.000,0.931} 56.6 & \cellcolor[rgb]{1.000,0.987,0.987} 48.6 & \cellcolor[rgb]{0.935,1.000,0.935} 54.9 & \cellcolor[rgb]{0.861,1.000,0.861} \textbf{56.8} \\
 & 24B tok & B2 simult. & \cellcolor[rgb]{0.896,1.000,0.896} 60.0 & \cellcolor[rgb]{0.976,1.000,0.976} 51.9 & \cellcolor[rgb]{0.895,1.000,0.895} 57.9 & \cellcolor[rgb]{0.969,1.000,0.969} 51.5 \\
 & 24B tok & B3 sequential & \cellcolor[rgb]{0.874,1.000,0.874} \textbf{62.1} & \cellcolor[rgb]{0.972,1.000,0.972} \textbf{52.2} & \cellcolor[rgb]{0.893,1.000,0.893} 58.1 & \cellcolor[rgb]{0.998,1.000,0.998} 50.1 \\
\bottomrule
\end{tabular}
\end{minipage}\hfill
\begin{minipage}[t]{0.49\linewidth}\centering
\begin{tabular}{ll l cccc}
\toprule
& & & \multicolumn{4}{c}{\textbf{BLiSS (matched-L1 cohort)}} \\
\cmidrule(lr){4-7}
\textbf{Pair} & \textbf{Data} & \textbf{Curr.} & \textbf{RP@0} & \textbf{RP@$\tau$} & \textbf{NGS} & \textbf{CPS} \\
\midrule
\texttt{zho--eng} & Human-scale & Mono & \cellcolor[rgb]{1.000,0.730,0.730} 45.4 & \cellcolor[rgb]{1.000,0.730,0.730} 22.0 & \cellcolor[rgb]{1.000,0.730,0.730} 46.5 & \cellcolor[rgb]{1.000,0.730,0.730} 25.5 \\
 & Human-scale & B2 simult. & \cellcolor[rgb]{0.856,1.000,0.856} 63.8 & \cellcolor[rgb]{0.942,1.000,0.942} 54.6 & \cellcolor[rgb]{0.857,1.000,0.857} 60.8 & \cellcolor[rgb]{1.000,0.851,0.851} 36.5 \\
 & Human-scale & B3 sequential & \cellcolor[rgb]{0.854,1.000,0.854} 64.0 & \cellcolor[rgb]{0.958,1.000,0.958} 53.3 & \cellcolor[rgb]{0.848,1.000,0.848} 61.5 & \cellcolor[rgb]{1.000,0.850,0.850} 36.4 \\
 & Human-scale & B5 late & \cellcolor[rgb]{0.851,1.000,0.851} 64.3 & \cellcolor[rgb]{0.937,1.000,0.937} 55.0 & \cellcolor[rgb]{0.843,1.000,0.843} 61.9 & \cellcolor[rgb]{1.000,0.850,0.850} 36.4 \\
\addlinespace[1.5pt]
 & 2B tok & B1 balanced & \cellcolor[rgb]{0.807,1.000,0.807} 68.5 & \cellcolor[rgb]{0.873,1.000,0.873} 60.1 & \cellcolor[rgb]{0.807,1.000,0.807} 64.6 & \cellcolor[rgb]{1.000,0.994,0.994} 49.5 \\
 & 2B tok & B2 simult. & \cellcolor[rgb]{0.808,1.000,0.808} 68.4 & \cellcolor[rgb]{0.886,1.000,0.886} 59.0 & \cellcolor[rgb]{0.813,1.000,0.813} 64.1 & \cellcolor[rgb]{1.000,0.983,0.983} 48.5 \\
 & 2B tok & B3 sequential & \cellcolor[rgb]{0.801,1.000,0.801} 69.1 & \cellcolor[rgb]{0.871,1.000,0.871} 60.2 & \cellcolor[rgb]{0.805,1.000,0.805} 64.7 & \cellcolor[rgb]{1.000,0.989,0.989} 49.0 \\
 & 2B tok & B4 classroom & \cellcolor[rgb]{0.846,1.000,0.846} 64.8 & \cellcolor[rgb]{0.923,1.000,0.923} 56.1 & \cellcolor[rgb]{0.823,1.000,0.823} 63.4 & \cellcolor[rgb]{1.000,0.905,0.905} 41.4 \\
 & 2B tok & B5 late & \cellcolor[rgb]{0.817,1.000,0.817} 67.6 & \cellcolor[rgb]{0.900,1.000,0.900} 57.9 & \cellcolor[rgb]{0.808,1.000,0.808} 64.5 & \cellcolor[rgb]{1.000,0.949,0.949} 45.4 \\
 \addlinespace[1.5pt]
 & 24B tok (FW-24B) & B1 balanced & \cellcolor[rgb]{0.796,1.000,0.796} 69.8 & \cellcolor[rgb]{0.856,1.000,0.856} 61.7 & \cellcolor[rgb]{0.804,1.000,0.804} 64.8 & \cellcolor[rgb]{1.000,0.932,0.932} 54.8 \\
 & 24B tok (FW-24B) & B2 simult. & \cellcolor[rgb]{0.792,1.000,0.792} 70.2 & \cellcolor[rgb]{0.849,1.000,0.849} 62.4 & \cellcolor[rgb]{0.804,1.000,0.804} 64.8 & \cellcolor[rgb]{1.000,0.934,0.934} 54.6 \\
 & 24B tok (FW-24B) & B3 sequential & \cellcolor[rgb]{0.811,1.000,0.811} 68.3 & \cellcolor[rgb]{0.859,1.000,0.859} 61.2 & \cellcolor[rgb]{0.816,1.000,0.816} 63.6 & \cellcolor[rgb]{1.000,0.928,0.928} 55.2 \\
 & 24B tok (FW-24B) & B3 sequential {\scriptsize\textit{ewc}} & \cellcolor[rgb]{0.796,1.000,0.796} 69.8 & \cellcolor[rgb]{0.858,1.000,0.858} 61.3 & \cellcolor[rgb]{0.810,1.000,0.810} 64.2 & \cellcolor[rgb]{1.000,0.919,0.919} \textbf{56.1} \\
 & 24B tok (FW-24B) & B4 classroom & \cellcolor[rgb]{0.804,1.000,0.804} 69.0 & \cellcolor[rgb]{0.852,1.000,0.852} 62.1 & \cellcolor[rgb]{0.805,1.000,0.805} 64.7 & \cellcolor[rgb]{1.000,0.961,0.961} 51.9 \\
 & 24B tok (FW-24B) & B5 late & \cellcolor[rgb]{0.787,1.000,0.787} \textbf{70.7} & \cellcolor[rgb]{0.843,1.000,0.843} \textbf{63.3} & \cellcolor[rgb]{0.802,1.000,0.802} \textbf{65.0} & \cellcolor[rgb]{1.000,0.942,0.942} 53.8 \\
\midrule
\midrule
\texttt{hin--eng} & 100M tok & B1 balanced & \cellcolor[rgb]{0.950,1.000,0.950} \textbf{54.8} & \cellcolor[rgb]{1.000,0.945,0.945} \textbf{44.3} & \cellcolor[rgb]{0.943,1.000,0.943} \textbf{54.3} & \cellcolor[rgb]{0.922,1.000,0.922} \textbf{53.8} \\
 & 100M tok & B2 simult. & \cellcolor[rgb]{0.975,1.000,0.975} 52.4 & \cellcolor[rgb]{1.000,0.910,0.910} 40.7 & \cellcolor[rgb]{0.952,1.000,0.952} 53.6 & \cellcolor[rgb]{1.000,0.917,0.917} 42.5 \\
 & 100M tok & B3 sequential & \cellcolor[rgb]{0.978,1.000,0.978} 52.1 & \cellcolor[rgb]{1.000,0.913,0.913} 41.0 & \cellcolor[rgb]{0.952,1.000,0.952} 53.6 & \cellcolor[rgb]{1.000,0.936,0.936} 44.2 \\
 & 100M tok & B4 classroom & \cellcolor[rgb]{0.955,1.000,0.955} 54.3 & \cellcolor[rgb]{1.000,0.918,0.918} 41.5 & \cellcolor[rgb]{0.948,1.000,0.948} 53.9 & \cellcolor[rgb]{1.000,0.964,0.964} 46.7 \\
 & 100M tok & B5 late & \cellcolor[rgb]{0.953,1.000,0.953} 54.5 & \cellcolor[rgb]{1.000,0.939,0.939} 43.7 & \cellcolor[rgb]{0.951,1.000,0.951} 53.7 & \cellcolor[rgb]{0.939,1.000,0.939} 53.0 \\
\addlinespace[1.5pt]
 & 2B tok & B2 simult. & \cellcolor[rgb]{0.969,1.000,0.969} 53.0 & \cellcolor[rgb]{1.000,0.910,0.910} 40.7 & \cellcolor[rgb]{0.960,1.000,0.960} 53.0 & \cellcolor[rgb]{1.000,0.910,0.910} 41.8 \\
\midrule
\texttt{ita--eng} & 100M tok & B1 balanced & \cellcolor[rgb]{0.869,1.000,0.869} \textbf{62.6} & \cellcolor[rgb]{0.984,1.000,0.984} \textbf{51.3} & \cellcolor[rgb]{0.872,1.000,0.872} 59.7 & \cellcolor[rgb]{1.000,0.992,0.992} \textbf{49.3} \\
 & 100M tok & B2 simult. & \cellcolor[rgb]{0.877,1.000,0.877} 61.8 & \cellcolor[rgb]{1.000,0.994,0.994} 49.4 & \cellcolor[rgb]{0.857,1.000,0.857} \textbf{60.8} & \cellcolor[rgb]{1.000,0.869,0.869} 38.1 \\
 & 100M tok & B3 sequential & \cellcolor[rgb]{0.874,1.000,0.874} 62.1 & \cellcolor[rgb]{1.000,0.986,0.986} 48.5 & \cellcolor[rgb]{0.870,1.000,0.870} 59.8 & \cellcolor[rgb]{1.000,0.871,0.871} 38.3 \\
 & 100M tok & B4 classroom & \cellcolor[rgb]{0.869,1.000,0.869} \textbf{62.6} & \cellcolor[rgb]{0.996,1.000,0.996} 50.3 & \cellcolor[rgb]{0.858,1.000,0.858} 60.7 & \cellcolor[rgb]{1.000,0.928,0.928} 43.5 \\
 & 100M tok & B5 late & \cellcolor[rgb]{0.880,1.000,0.880} 61.5 & \cellcolor[rgb]{0.992,1.000,0.992} 50.6 & \cellcolor[rgb]{0.873,1.000,0.873} 59.6 & \cellcolor[rgb]{1.000,0.972,0.972} 47.5 \\
\addlinespace[1.5pt]
 & 2B tok & B3 sequential & \cellcolor[rgb]{0.883,1.000,0.883} 61.2 & \cellcolor[rgb]{0.996,1.000,0.996} 50.3 & \cellcolor[rgb]{0.884,1.000,0.884} 58.8 & \cellcolor[rgb]{1.000,0.979,0.979} 48.1 \\
\midrule
\texttt{pol--eng} & 100M tok & B1 balanced & \cellcolor[rgb]{0.948,1.000,0.948} \textbf{55.0} & \cellcolor[rgb]{1.000,0.972,0.972} \textbf{47.1} & \cellcolor[rgb]{0.943,1.000,0.943} \textbf{54.3} & \cellcolor[rgb]{0.906,1.000,0.906} \textbf{54.6} \\
 & 100M tok & B2 simult. & \cellcolor[rgb]{0.977,1.000,0.977} 52.2 & \cellcolor[rgb]{1.000,0.941,0.941} 43.9 & \cellcolor[rgb]{0.959,1.000,0.959} 53.1 & \cellcolor[rgb]{1.000,0.946,0.946} 45.1 \\
 & 100M tok & B3 sequential & \cellcolor[rgb]{0.978,1.000,0.978} 52.1 & \cellcolor[rgb]{1.000,0.941,0.941} 43.9 & \cellcolor[rgb]{0.963,1.000,0.963} 52.8 & \cellcolor[rgb]{1.000,0.960,0.960} 46.4 \\
\midrule
\texttt{tur--eng} & 100M tok & B1 balanced & \cellcolor[rgb]{0.922,1.000,0.922} \textbf{57.5} & \cellcolor[rgb]{0.997,1.000,0.997} \textbf{50.2} & \cellcolor[rgb]{0.925,1.000,0.925} \textbf{55.7} & \cellcolor[rgb]{0.988,1.000,0.988} \textbf{50.6} \\
 & 100M tok & B2 simult. & \cellcolor[rgb]{0.967,1.000,0.967} 53.2 & \cellcolor[rgb]{1.000,0.951,0.951} 44.9 & \cellcolor[rgb]{0.956,1.000,0.956} 53.3 & \cellcolor[rgb]{1.000,0.888,0.888} 39.8 \\
 & 100M tok & B3 sequential & \cellcolor[rgb]{0.971,1.000,0.971} 52.8 & \cellcolor[rgb]{1.000,0.950,0.950} 44.8 & \cellcolor[rgb]{0.958,1.000,0.958} 53.2 & \cellcolor[rgb]{1.000,0.881,0.881} 39.2 \\
 & 100M tok & B4 classroom & \cellcolor[rgb]{0.934,1.000,0.934} 56.3 & \cellcolor[rgb]{1.000,0.984,0.984} 48.3 & \cellcolor[rgb]{0.926,1.000,0.926} 55.6 & \cellcolor[rgb]{1.000,0.923,0.923} 43.0 \\
 & 100M tok & B5 late & \cellcolor[rgb]{0.936,1.000,0.936} 56.1 & \cellcolor[rgb]{1.000,0.991,0.991} 49.1 & \cellcolor[rgb]{0.929,1.000,0.929} 55.4 & \cellcolor[rgb]{1.000,0.986,0.986} 48.7 \\
\midrule
\texttt{eng--deu} & 24B tok & B1 balanced & \cellcolor[rgb]{0.952,1.000,0.952} \textbf{54.6} & \cellcolor[rgb]{1.000,0.964,0.964} \textbf{46.3} & \cellcolor[rgb]{0.962,1.000,0.962} \textbf{52.9} & \cellcolor[rgb]{0.963,1.000,0.963} \textbf{51.8} \\
\midrule
\texttt{eng--nld} & 24B tok & B1 balanced & \cellcolor[rgb]{0.929,1.000,0.929} \textbf{56.8} & \cellcolor[rgb]{0.997,1.000,0.997} \textbf{50.2} & \cellcolor[rgb]{0.938,1.000,0.938} \textbf{54.7} & \cellcolor[rgb]{0.904,1.000,0.904} 54.7 \\
 & 24B tok & B2 simult. & \cellcolor[rgb]{0.931,1.000,0.931} 56.6 & \cellcolor[rgb]{1.000,0.987,0.987} 48.6 & \cellcolor[rgb]{0.946,1.000,0.946} 54.1 & \cellcolor[rgb]{0.904,1.000,0.904} 54.7 \\
 & 24B tok & B3 sequential & \cellcolor[rgb]{0.952,1.000,0.952} 54.6 & \cellcolor[rgb]{1.000,0.979,0.979} 47.8 & \cellcolor[rgb]{0.955,1.000,0.955} 53.4 & \cellcolor[rgb]{0.881,1.000,0.881} \textbf{55.8} \\
\bottomrule
\end{tabular}
\end{minipage}
\caption{BLiSS learner-minimal-pair scores (\%, 0--100) for \beetle{} models on the \emph{matched-L1 cohort}, grouped by L1 then token budget. Cell shading encodes score on a fixed scale (darker green = higher). Per L1, the highest value in each metric column is shown in \textbf{bold}. Metrics: RP@0 / RP@$\tau$ (random-over-human preference, with and without a margin), NGS (normalised gap score), CPS (correct-preferred sanity check).}
\label{tab:bliss_beetle}
\end{table*}

\begin{table}[!ht]
\centering\scriptsize
\setlength{\tabcolsep}{3pt}
\begin{tabular}{ll l cccc}
\toprule
& & & \multicolumn{4}{c}{\textbf{BLiSS (matched-L1 cohort)}} \\
\cmidrule(lr){4-7}
\textbf{Model} & \textbf{Params} & \textbf{L1} & \textbf{RP@0} & \textbf{RP@$\tau$} & \textbf{NGS} & \textbf{CPS} \\
\midrule
BLOOM-560M & 0.56B & Dutch & \cellcolor[rgb]{0.906,1.000,0.906} 59.0 & \cellcolor[rgb]{0.966,1.000,0.966} 52.7 & \cellcolor[rgb]{0.929,1.000,0.929} 55.4 & \cellcolor[rgb]{0.881,1.000,0.881} 55.8 \\
 &  & German & \cellcolor[rgb]{0.907,1.000,0.907} 58.9 & \cellcolor[rgb]{0.991,1.000,0.991} 50.7 & \cellcolor[rgb]{0.935,1.000,0.935} 54.9 & \cellcolor[rgb]{0.953,1.000,0.953} 52.3 \\
 &  & Chinese & \cellcolor[rgb]{0.789,1.000,0.789} \textbf{70.2} & \cellcolor[rgb]{0.847,1.000,0.847} \textbf{62.1} & \cellcolor[rgb]{0.801,1.000,0.801} \textbf{65.0} & \cellcolor[rgb]{0.875,1.000,0.875} \textbf{56.1} \\
\addlinespace[2pt]
BLOOM-1.1B & 1.1B & Dutch & \cellcolor[rgb]{0.899,1.000,0.899} 59.7 & \cellcolor[rgb]{0.980,1.000,0.980} 51.6 & \cellcolor[rgb]{0.923,1.000,0.923} 55.8 & \cellcolor[rgb]{0.865,1.000,0.865} \textbf{56.6} \\
 &  & German & \cellcolor[rgb]{0.921,1.000,0.921} 57.6 & \cellcolor[rgb]{1.000,0.994,0.994} 49.4 & \cellcolor[rgb]{0.938,1.000,0.938} 54.7 & \cellcolor[rgb]{0.937,1.000,0.937} 53.1 \\
 &  & Chinese & \cellcolor[rgb]{0.788,1.000,0.788} \textbf{70.3} & \cellcolor[rgb]{0.828,1.000,0.828} \textbf{63.6} & \cellcolor[rgb]{0.792,1.000,0.792} \textbf{65.7} & \cellcolor[rgb]{0.865,1.000,0.865} \textbf{56.6} \\
\addlinespace[2pt]
BLOOM-1.7B & 1.7B & Dutch & \cellcolor[rgb]{0.906,1.000,0.906} 59.0 & \cellcolor[rgb]{0.984,1.000,0.984} 51.3 & \cellcolor[rgb]{0.922,1.000,0.922} 55.9 & \cellcolor[rgb]{0.865,1.000,0.865} 56.6 \\
 &  & German & \cellcolor[rgb]{0.924,1.000,0.924} 57.3 & \cellcolor[rgb]{0.997,1.000,0.997} 50.2 & \cellcolor[rgb]{0.938,1.000,0.938} 54.7 & \cellcolor[rgb]{0.883,1.000,0.883} 55.7 \\
 &  & Chinese & \cellcolor[rgb]{0.786,1.000,0.786} \textbf{70.5} & \cellcolor[rgb]{0.827,1.000,0.827} \textbf{63.7} & \cellcolor[rgb]{0.792,1.000,0.792} \textbf{65.7} & \cellcolor[rgb]{0.828,1.000,0.828} \textbf{58.4} \\
\addlinespace[2pt]
BLOOM-7.1B & 7.1B & Dutch & \cellcolor[rgb]{0.863,1.000,0.863} 63.1 & \cellcolor[rgb]{0.909,1.000,0.909} 57.2 & \cellcolor[rgb]{0.890,1.000,0.890} 58.3 & \cellcolor[rgb]{0.820,1.000,0.820} \textbf{58.8} \\
 &  & German & \cellcolor[rgb]{0.903,1.000,0.903} 59.3 & \cellcolor[rgb]{0.999,1.000,0.999} 50.0 & \cellcolor[rgb]{0.925,1.000,0.925} 55.7 & \cellcolor[rgb]{0.863,1.000,0.863} 56.7 \\
 &  & Chinese & \cellcolor[rgb]{0.758,1.000,0.758} \textbf{73.2} & \cellcolor[rgb]{0.803,1.000,0.803} \textbf{65.6} & \cellcolor[rgb]{0.770,1.000,0.770} \textbf{67.4} & \cellcolor[rgb]{0.847,1.000,0.847} 57.5 \\
\addlinespace[2pt]
XGLM-4.5B & 4.5B & Dutch & \cellcolor[rgb]{0.875,1.000,0.875} 62.0 & \cellcolor[rgb]{0.950,1.000,0.950} 54.0 & \cellcolor[rgb]{0.901,1.000,0.901} 57.5 & \cellcolor[rgb]{0.730,1.000,0.730} \textbf{63.2} \\
 &  & German & \cellcolor[rgb]{0.903,1.000,0.903} 59.3 & \cellcolor[rgb]{0.992,1.000,0.992} 50.6 & \cellcolor[rgb]{0.930,1.000,0.930} 55.3 & \cellcolor[rgb]{0.834,1.000,0.834} 58.1 \\
 &  & Chinese & \cellcolor[rgb]{0.753,1.000,0.753} \textbf{73.7} & \cellcolor[rgb]{0.802,1.000,0.802} \textbf{65.7} & \cellcolor[rgb]{0.771,1.000,0.771} \textbf{67.3} & \cellcolor[rgb]{0.800,1.000,0.800} 59.8 \\
\addlinespace[2pt]
EuroLLM-9B & 9B & Dutch & \cellcolor[rgb]{0.850,1.000,0.850} 64.4 & \cellcolor[rgb]{0.921,1.000,0.921} 56.3 & \cellcolor[rgb]{0.901,1.000,0.901} 57.5 & \cellcolor[rgb]{0.763,1.000,0.763} \textbf{61.6} \\
 &  & German & \cellcolor[rgb]{0.893,1.000,0.893} 60.3 & \cellcolor[rgb]{0.990,1.000,0.990} 50.8 & \cellcolor[rgb]{0.932,1.000,0.932} 55.1 & \cellcolor[rgb]{0.769,1.000,0.769} 61.3 \\
 &  & Chinese & \cellcolor[rgb]{0.753,1.000,0.753} \textbf{73.7} & \cellcolor[rgb]{0.806,1.000,0.806} \textbf{65.4} & \cellcolor[rgb]{0.768,1.000,0.768} \textbf{67.5} & \cellcolor[rgb]{0.773,1.000,0.773} 61.1 \\
\addlinespace[2pt]
Gemma-3-270M & 0.27B & Dutch & \cellcolor[rgb]{0.887,1.000,0.887} 60.8 & \cellcolor[rgb]{0.936,1.000,0.936} 55.1 & \cellcolor[rgb]{0.905,1.000,0.905} 57.2 & \cellcolor[rgb]{0.859,1.000,0.859} \textbf{56.9} \\
 &  & German & \cellcolor[rgb]{0.925,1.000,0.925} 57.2 & \cellcolor[rgb]{0.976,1.000,0.976} 51.9 & \cellcolor[rgb]{0.930,1.000,0.930} 55.3 & \cellcolor[rgb]{0.926,1.000,0.926} 53.6 \\
 &  & Chinese & \cellcolor[rgb]{0.771,1.000,0.771} \textbf{72.0} & \cellcolor[rgb]{0.792,1.000,0.792} \textbf{66.5} & \cellcolor[rgb]{0.779,1.000,0.779} \textbf{66.7} & \cellcolor[rgb]{0.875,1.000,0.875} 56.1 \\
\addlinespace[2pt]
Gemma-2-2B & 2B & Dutch & \cellcolor[rgb]{0.847,1.000,0.847} 64.7 & \cellcolor[rgb]{0.907,1.000,0.907} 57.4 & \cellcolor[rgb]{0.893,1.000,0.893} 58.1 & \cellcolor[rgb]{0.875,1.000,0.875} \textbf{56.1} \\
 &  & German & \cellcolor[rgb]{0.864,1.000,0.864} 63.0 & \cellcolor[rgb]{0.914,1.000,0.914} 56.8 & \cellcolor[rgb]{0.887,1.000,0.887} 58.5 & \cellcolor[rgb]{0.894,1.000,0.894} 55.2 \\
 &  & Chinese & \cellcolor[rgb]{0.730,1.000,0.730} \textbf{75.9} & \cellcolor[rgb]{0.731,1.000,0.731} \textbf{71.3} & \cellcolor[rgb]{0.741,1.000,0.741} \textbf{69.6} & \cellcolor[rgb]{0.908,1.000,0.908} 54.5 \\
\addlinespace[2pt]
Gemma-2-9B & 9B & Dutch & \cellcolor[rgb]{0.838,1.000,0.838} 65.5 & \cellcolor[rgb]{0.869,1.000,0.869} 60.4 & \cellcolor[rgb]{0.872,1.000,0.872} 59.7 & \cellcolor[rgb]{0.908,1.000,0.908} 54.5 \\
 &  & German & \cellcolor[rgb]{0.833,1.000,0.833} 66.0 & \cellcolor[rgb]{0.869,1.000,0.869} 60.4 & \cellcolor[rgb]{0.868,1.000,0.868} 60.0 & \cellcolor[rgb]{0.890,1.000,0.890} \textbf{55.4} \\
 &  & Chinese & \cellcolor[rgb]{0.736,1.000,0.736} \textbf{75.3} & \cellcolor[rgb]{0.730,1.000,0.730} \textbf{71.4} & \cellcolor[rgb]{0.730,1.000,0.730} \textbf{70.4} & \cellcolor[rgb]{0.894,1.000,0.894} 55.2 \\
\addlinespace[2pt]
Gemma-2-27B & 27B & Dutch & \cellcolor[rgb]{0.802,1.000,0.802} 69.0 & \cellcolor[rgb]{0.857,1.000,0.857} 61.3 & \cellcolor[rgb]{0.828,1.000,0.828} 63.0 & \cellcolor[rgb]{0.883,1.000,0.883} \textbf{55.7} \\
 &  & German & \cellcolor[rgb]{0.811,1.000,0.811} 68.1 & \cellcolor[rgb]{0.883,1.000,0.883} 59.3 & \cellcolor[rgb]{0.841,1.000,0.841} 62.0 & \cellcolor[rgb]{0.957,1.000,0.957} 52.1 \\
 &  & Chinese & \cellcolor[rgb]{0.737,1.000,0.737} \textbf{75.2} & \cellcolor[rgb]{0.748,1.000,0.748} \textbf{70.0} & \cellcolor[rgb]{0.731,1.000,0.731} \textbf{70.3} & \cellcolor[rgb]{0.914,1.000,0.914} 54.2 \\
\addlinespace[2pt]
Qwen3-0.6B & 0.6B & Dutch & \cellcolor[rgb]{0.901,1.000,0.901} 59.5 & \cellcolor[rgb]{0.948,1.000,0.948} 54.1 & \cellcolor[rgb]{0.925,1.000,0.925} 55.7 & \cellcolor[rgb]{0.789,1.000,0.789} 60.3 \\
 &  & German & \cellcolor[rgb]{0.919,1.000,0.919} 57.8 & \cellcolor[rgb]{1.000,0.999,0.999} 49.9 & \cellcolor[rgb]{0.929,1.000,0.929} 55.4 & \cellcolor[rgb]{0.849,1.000,0.849} 57.4 \\
 &  & Chinese & \cellcolor[rgb]{0.782,1.000,0.782} \textbf{70.9} & \cellcolor[rgb]{0.846,1.000,0.846} \textbf{62.2} & \cellcolor[rgb]{0.796,1.000,0.796} \textbf{65.4} & \cellcolor[rgb]{0.777,1.000,0.777} \textbf{60.9} \\
\addlinespace[2pt]
Qwen3-4B & 4B & Dutch & \cellcolor[rgb]{0.862,1.000,0.862} 63.2 & \cellcolor[rgb]{0.928,1.000,0.928} 55.7 & \cellcolor[rgb]{0.882,1.000,0.882} 58.9 & \cellcolor[rgb]{0.771,1.000,0.771} \textbf{61.2} \\
 &  & German & \cellcolor[rgb]{0.893,1.000,0.893} 60.3 & \cellcolor[rgb]{0.957,1.000,0.957} 53.4 & \cellcolor[rgb]{0.925,1.000,0.925} 55.7 & \cellcolor[rgb]{0.808,1.000,0.808} 59.4 \\
 &  & Chinese & \cellcolor[rgb]{0.762,1.000,0.762} \textbf{72.8} & \cellcolor[rgb]{0.797,1.000,0.797} \textbf{66.1} & \cellcolor[rgb]{0.771,1.000,0.771} \textbf{67.3} & \cellcolor[rgb]{0.871,1.000,0.871} 56.3 \\
\addlinespace[2pt]
Qwen3-8B & 8B & Dutch & \cellcolor[rgb]{0.839,1.000,0.839} 65.4 & \cellcolor[rgb]{0.912,1.000,0.912} 57.0 & \cellcolor[rgb]{0.874,1.000,0.874} 59.5 & \cellcolor[rgb]{0.755,1.000,0.755} \textbf{62.0} \\
 &  & German & \cellcolor[rgb]{0.897,1.000,0.897} 59.9 & \cellcolor[rgb]{0.967,1.000,0.967} 52.6 & \cellcolor[rgb]{0.917,1.000,0.917} 56.3 & \cellcolor[rgb]{0.845,1.000,0.845} 57.6 \\
 &  & Chinese & \cellcolor[rgb]{0.737,1.000,0.737} \textbf{75.2} & \cellcolor[rgb]{0.777,1.000,0.777} \textbf{67.7} & \cellcolor[rgb]{0.750,1.000,0.750} \textbf{68.9} & \cellcolor[rgb]{0.843,1.000,0.843} 57.7 \\
\addlinespace[2pt]
Llama-3.2-1B & 1B & Dutch & \cellcolor[rgb]{0.859,1.000,0.859} 63.5 & \cellcolor[rgb]{0.938,1.000,0.938} 54.9 & \cellcolor[rgb]{0.895,1.000,0.895} 57.9 & \cellcolor[rgb]{0.779,1.000,0.779} \textbf{60.8} \\
 &  & German & \cellcolor[rgb]{0.883,1.000,0.883} 61.2 & \cellcolor[rgb]{0.956,1.000,0.956} 53.5 & \cellcolor[rgb]{0.914,1.000,0.914} 56.5 & \cellcolor[rgb]{0.812,1.000,0.812} 59.2 \\
 &  & Chinese & \cellcolor[rgb]{0.751,1.000,0.751} \textbf{73.9} & \cellcolor[rgb]{0.797,1.000,0.797} \textbf{66.1} & \cellcolor[rgb]{0.766,1.000,0.766} \textbf{67.7} & \cellcolor[rgb]{0.793,1.000,0.793} 60.1 \\
\addlinespace[2pt]
Llama-3.1-8B & 8B & Dutch & \cellcolor[rgb]{0.823,1.000,0.823} 67.0 & \cellcolor[rgb]{0.890,1.000,0.890} 58.7 & \cellcolor[rgb]{0.858,1.000,0.858} 60.7 & \cellcolor[rgb]{0.783,1.000,0.783} \textbf{60.6} \\
 &  & German & \cellcolor[rgb]{0.871,1.000,0.871} 62.4 & \cellcolor[rgb]{0.939,1.000,0.939} 54.8 & \cellcolor[rgb]{0.894,1.000,0.894} 58.0 & \cellcolor[rgb]{0.851,1.000,0.851} 57.3 \\
 &  & Chinese & \cellcolor[rgb]{0.731,1.000,0.731} \textbf{75.8} & \cellcolor[rgb]{0.764,1.000,0.764} \textbf{68.7} & \cellcolor[rgb]{0.746,1.000,0.746} \textbf{69.2} & \cellcolor[rgb]{0.810,1.000,0.810} 59.3 \\
\bottomrule
\end{tabular}
\caption{BLiSS learner-minimal-pair scores (\%, 0--100) for frontier
multilingual-LLM baselines, scored per L1 cohort. \emph{Params} gives the
parameter count and \emph{L1} the cohort. Cell shading encodes score on a
fixed scale (darker green = higher). Per model, the highest value in each
metric column is in \textbf{bold}. Metrics: RP@0 / RP@$\tau$
(random-over-human preference, with and without a margin), NGS (normalised
gap score), CPS (correct-preferred sanity check).}
\label{tab:bliss_baselines}
\end{table}

\clearpage

\begin{table*}[!ht]
  \centering \scriptsize
  \setlength{\tabcolsep}{4pt}
  \caption{\textbf{Downstream pedagogical performance of \beetle{} models.} CLOTH closed-cloze accuracy, Cambridge open-cloze accuracy, JFLEG learner-fluency margin (bits/token; positive = model prefers corrected over learner-error sentence), false-friend disambiguation margin (bits/token), CEFR-A1 / CEFR-C2 BPT, and CEFR gradient. Bold = best per column within the matched-L1 nld--eng family.}
  \label{tab:headline_pedagogical}
  \begin{tabular}{llrrrrrrr}
    \toprule
    Model & Family & CLOTH & Camb.~OC & JFLEG & FF & A1 BPT & C2 BPT & A1$\to$C2 \\
    \midrule
    B1 nld--eng     & HS & \textbf{0.380} & \textbf{0.644} & \textbf{0.949} & \textbf{0.585} & 6.572 & 8.818 & 2.247 \\
    B2 nld--eng     & HS & 0.302 & 0.596 & 0.750 & 0.156 & 8.051 & 10.213 & 2.163 \\
    B3-33 nld--eng  & HS & 0.310 & 0.587 & 0.751 & 0.174 & 8.106 & 10.275 & 2.169 \\
    B4 nld--eng     & HS & 0.304 & 0.582 & 0.801 & $-$0.055 & 7.719 & 9.755 & 2.036 \\
    B5 nld--eng     & HS & 0.302 & 0.584 & 0.716 & 0.238 & 8.127 & 10.193 & 2.066 \\
    \midrule
    B1 nld--eng     & FW-2B & 0.369 & 0.638 & 0.833 & 0.457 & 8.851 & 8.614 & $-$0.237 \\
    B2 nld--eng     & FW-2B & 0.322 & 0.567 & 0.502 & 0.238 & 10.317 & 11.047 & 0.730 \\
    B3-33 nld--eng  & FW-2B & 0.329 & 0.614 & 0.511 & 0.235 & 10.354 & 11.056 & 0.702 \\
    B5 nld--eng     & FW-2B & 0.323 & 0.570 & 0.509 & 0.318 & 10.364 & 11.040 & 0.676 \\
    \midrule
    mono-eng        & HS & 0.369 & 0.659 & 0.888 & 0.351 & 7.297 & 10.558 & \textbf{3.261} \\
    mono-nld        & HS & 0.267 & 0.310 & 0.306 & $-$0.034 & 9.875 & 10.671 & 0.796 \\
    \bottomrule
  \end{tabular}
\end{table*}
 
\clearpage

\newpage 
\section{Detailed \beetle{} BLiMP and MultiBLiMP results}\label{blimp}

\autoref{tab:beetle_main} shows BLiMP and MultiBLiMP performance of \beetle{} models by L1 and scale (\autoref{tab:beetle_main}). 

\begin{figure}[!ht]
    \centering
    \includegraphics[width=1\linewidth]{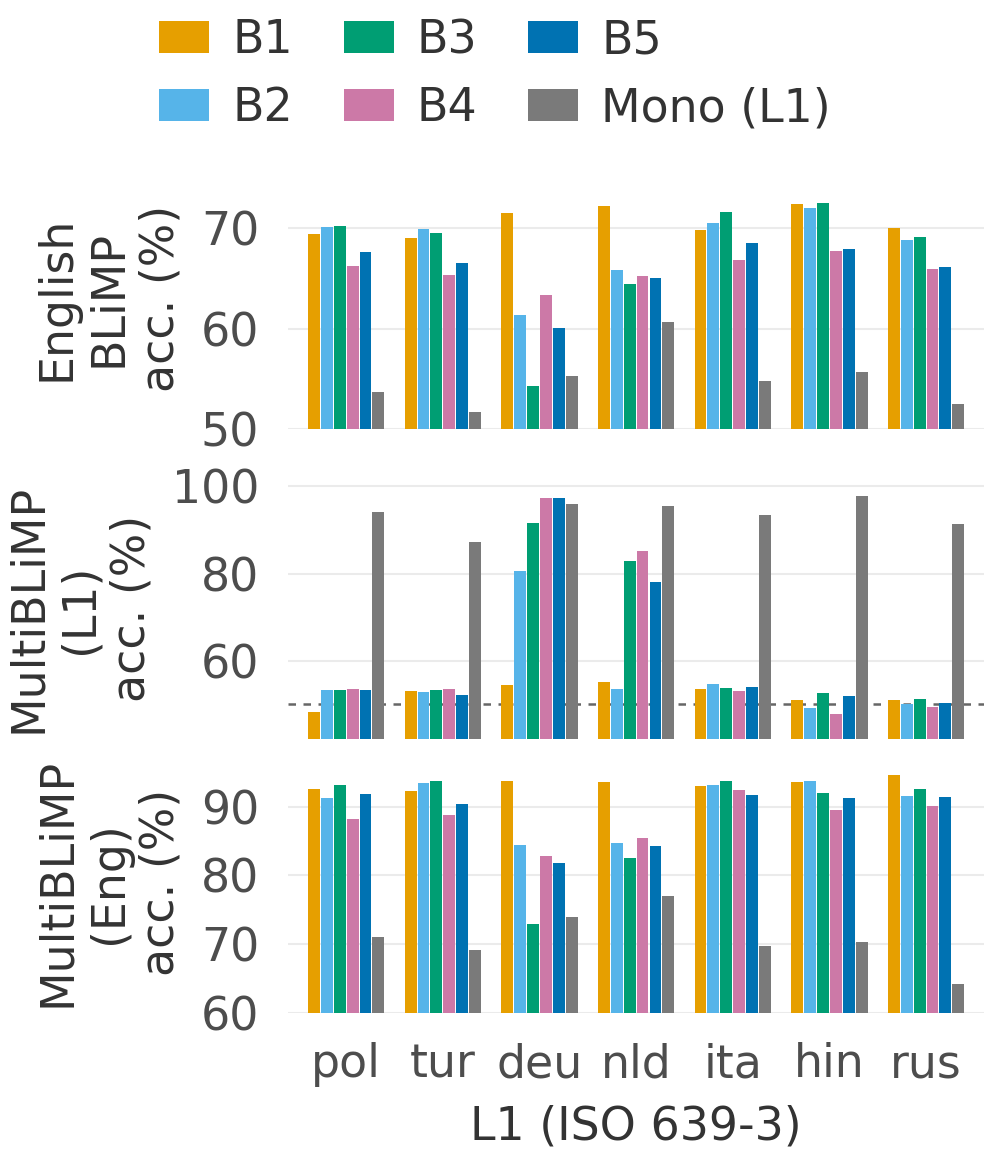}
    \caption{Per-phenomenon BLiMP accuracy for the 24B-token \beetle{} models, broken down by curriculum and L1. At this scale, differences between balanced and staged curricula are localised to specific phenomena rather than uniform across the benchmark, with the balanced and simultaneous curricula dominating on most syntactic and agreement phenomena.}
    \label{fig:blimp-grid-24b}
\end{figure}

\begin{figure}[!ht]
    \centering
    \includegraphics[width=1\linewidth]{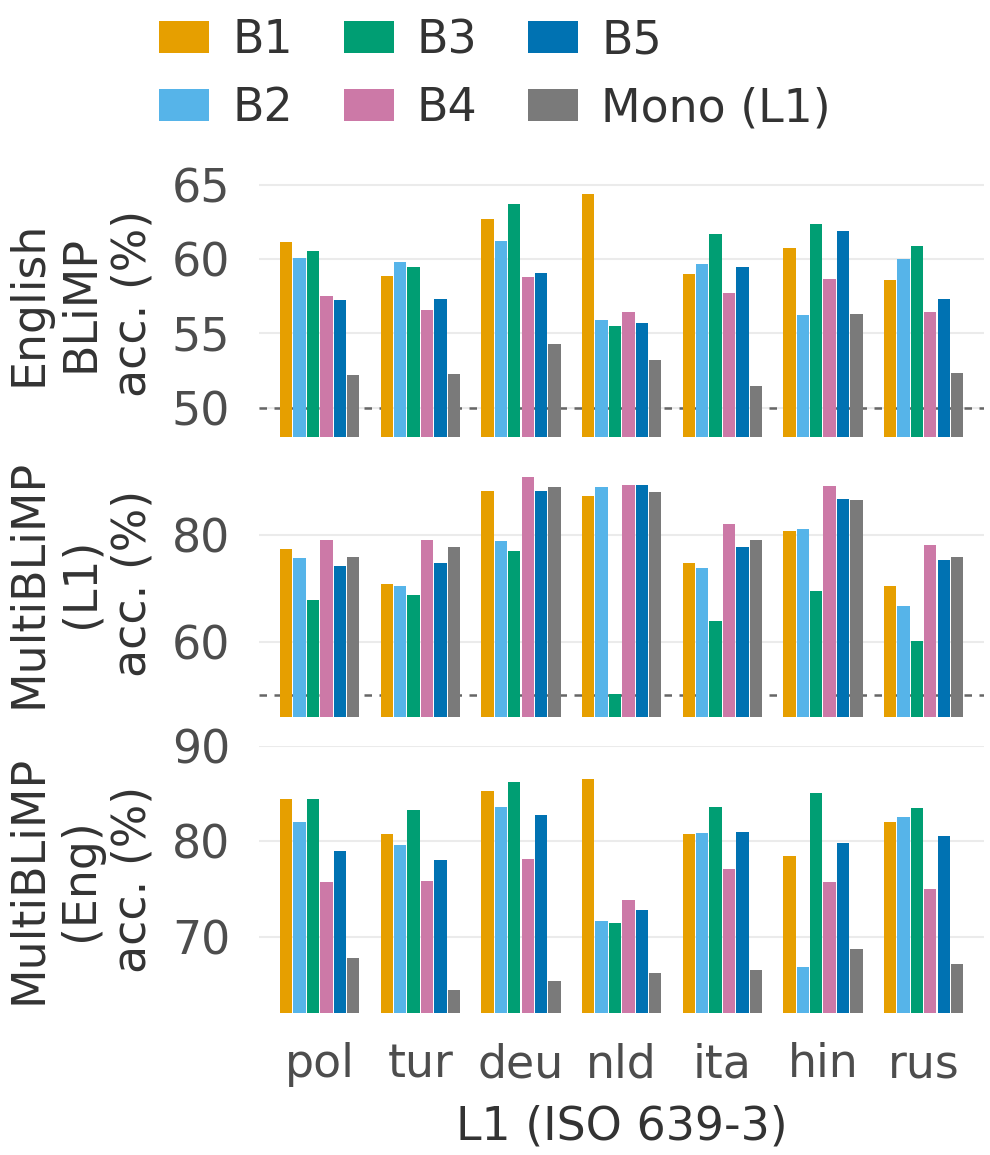}
    \caption{Per-phenomenon BLiMP accuracy for the 2B-token \beetle{} models, broken down by curriculum and L1. The overall curriculum ordering mirrors the 24B grid in \autoref{fig:blimp-grid-24b}, but accuracies are lower and the per-phenomenon gaps between curricula are wider, indicating that the additional data at 24B mainly narrows the harder phenomena rather than reshuffling which curriculum wins.}
    \label{fig:blimp-grid-2b}
\end{figure}

\autoref{fig:blimp-grid-24b} and \autoref{fig:blimp-grid-2b} give the per-phenomenon BLiMP accuracy grids for the 24B and 2B \beetle{} models, respectively, and \autoref{tab:baselines} reports frontier multilingual-LLM baselines for comparison.

\clearpage
\begin{table*}[ht!]
\centering\scriptsize
\setlength{\tabcolsep}{3pt}
\begin{minipage}[t]{0.49\linewidth}\centering
\begin{tabular}{ll l cc cc}
\toprule
& & & \multicolumn{2}{c}{\textbf{MultiBLiMP}} & \multicolumn{2}{c}{\textbf{MonoBLiMP}} \\
\cmidrule(lr){4-5}\cmidrule(lr){6-7}
\textbf{Pair} & \textbf{Data} & \textbf{Curr.} & \textbf{L1} & \textbf{Eng} & \textbf{L1} & \textbf{Eng} \\
\midrule
\texttt{deu--eng} & Human-scale & Mono & \cellcolor[rgb]{0.779,1.000,0.779} 89.9 & \cellcolor[rgb]{0.920,1.000,0.920} 64.7 & \textbf{-} & \cellcolor[rgb]{0.990,1.000,0.990} 51.2 \\
 & Human-scale & B1 balanced & \cellcolor[rgb]{0.796,1.000,0.796} 86.9 & \cellcolor[rgb]{0.834,1.000,0.834} 80.5 & \textbf{-} & \cellcolor[rgb]{0.872,1.000,0.872} 65.5 \\
 & Human-scale & B2 simult. & \cellcolor[rgb]{0.781,1.000,0.781} 89.5 & \cellcolor[rgb]{0.886,1.000,0.886} 70.9 & \textbf{-} & \cellcolor[rgb]{0.950,1.000,0.950} 56.1 \\
 & Human-scale & B3 sequential & \cellcolor[rgb]{0.779,1.000,0.779} 89.9 & \cellcolor[rgb]{0.884,1.000,0.884} 71.3 & \textbf{-} & \cellcolor[rgb]{0.939,1.000,0.939} 57.4 \\
 & Human-scale & B4 classroom & \cellcolor[rgb]{0.780,1.000,0.780} 89.7 & \cellcolor[rgb]{0.879,1.000,0.879} 72.2 & \textbf{-} & \cellcolor[rgb]{0.942,1.000,0.942} 57.0 \\
 & Human-scale & B5 late & \cellcolor[rgb]{0.781,1.000,0.781} 89.6 & \cellcolor[rgb]{0.884,1.000,0.884} 71.2 & \textbf{-} & \cellcolor[rgb]{0.952,1.000,0.952} 55.8 \\
\addlinespace[1.5pt]
 & 100M tok & B1 balanced & \cellcolor[rgb]{0.786,1.000,0.786} 88.6 & \cellcolor[rgb]{0.794,1.000,0.794} 87.7 & \textbf{-} & \cellcolor[rgb]{0.888,1.000,0.888} 63.6 \\
 & 100M tok & B2 simult. & \cellcolor[rgb]{0.767,1.000,0.767} 92.1 & \cellcolor[rgb]{0.884,1.000,0.884} 71.3 & \textbf{-} & \cellcolor[rgb]{0.977,1.000,0.977} 52.8 \\
 & 100M tok & B3 sequential & \cellcolor[rgb]{0.769,1.000,0.769} 91.7 & \cellcolor[rgb]{0.878,1.000,0.878} 72.3 & \textbf{-} & \cellcolor[rgb]{0.960,1.000,0.960} 54.8 \\
 & 100M tok & B4 classroom & \cellcolor[rgb]{0.792,1.000,0.792} 87.6 & \cellcolor[rgb]{0.860,1.000,0.860} 75.6 & \textbf{-} & \cellcolor[rgb]{0.942,1.000,0.942} 57.1 \\
 & 100M tok & B5 late & \cellcolor[rgb]{0.861,1.000,0.861} 75.2 & \cellcolor[rgb]{0.820,1.000,0.820} 83.0 & \textbf{-} & \cellcolor[rgb]{0.895,1.000,0.895} 62.7 \\
\addlinespace[1.5pt]
 & 2B tok & B1 balanced & \cellcolor[rgb]{0.789,1.000,0.789} 88.2 & \cellcolor[rgb]{0.807,1.000,0.807} 85.3 & \textbf{-} & \cellcolor[rgb]{0.895,1.000,0.895} 62.7 \\
 & 2B tok & B2 simult. & \cellcolor[rgb]{0.840,1.000,0.840} 78.9 & \cellcolor[rgb]{0.817,1.000,0.817} 83.6 & \textbf{-} & \cellcolor[rgb]{0.907,1.000,0.907} 61.3 \\
 & 2B tok & B3 sequential & \cellcolor[rgb]{0.851,1.000,0.851} 76.9 & \cellcolor[rgb]{0.803,1.000,0.803} 86.2 & \textbf{-} & \cellcolor[rgb]{0.887,1.000,0.887} 63.7 \\
 & 2B tok & B4 classroom & \cellcolor[rgb]{0.774,1.000,0.774} 90.8 & \cellcolor[rgb]{0.846,1.000,0.846} 78.2 & \textbf{-} & \cellcolor[rgb]{0.928,1.000,0.928} 58.8 \\
 & 2B tok & B5 late & \cellcolor[rgb]{0.788,1.000,0.788} 88.3 & \cellcolor[rgb]{0.822,1.000,0.822} 82.7 & \textbf{-} & \cellcolor[rgb]{0.925,1.000,0.925} 59.1 \\
\addlinespace[1.5pt]
 & 24B tok & B1 balanced & \cellcolor[rgb]{0.747,1.000,0.747} 95.7 & \cellcolor[rgb]{0.760,1.000,0.760} \textbf{94.0} & \textbf{-} & \cellcolor[rgb]{0.823,1.000,0.823} \textbf{71.5} \\
 & 24B tok & B2 simult. & \cellcolor[rgb]{0.742,1.000,0.742} \textbf{96.6} & \cellcolor[rgb]{0.812,1.000,0.812} 84.4 & \textbf{-} & \cellcolor[rgb]{0.907,1.000,0.907} 61.3 \\
\midrule
\texttt{eng} & Human-scale & Mono & \textbf{-} & \cellcolor[rgb]{0.842,1.000,0.842} \textbf{79.0} & \textbf{-} & \cellcolor[rgb]{0.879,1.000,0.879} \textbf{64.7} \\
\midrule
\texttt{nld--eng} & Human-scale & Mono & \cellcolor[rgb]{0.785,1.000,0.785} 88.9 & \cellcolor[rgb]{0.926,1.000,0.926} 63.5 & \cellcolor[rgb]{0.794,1.000,0.794} 79.0 & \cellcolor[rgb]{0.979,1.000,0.979} 52.5 \\
 & Human-scale & B1 balanced & \cellcolor[rgb]{0.792,1.000,0.792} 87.6 & \cellcolor[rgb]{0.832,1.000,0.832} 80.8 & \cellcolor[rgb]{0.810,1.000,0.810} 76.8 & \cellcolor[rgb]{0.843,1.000,0.843} 69.1 \\
 & Human-scale & B2 simult. & \cellcolor[rgb]{0.785,1.000,0.785} 88.8 & \cellcolor[rgb]{0.869,1.000,0.869} 74.0 & \cellcolor[rgb]{0.793,1.000,0.793} 79.1 & \cellcolor[rgb]{0.940,1.000,0.940} 57.3 \\
 & Human-scale & B3 sequential & \cellcolor[rgb]{0.777,1.000,0.777} 90.3 & \cellcolor[rgb]{0.870,1.000,0.870} 73.8 & \cellcolor[rgb]{0.799,1.000,0.799} 78.3 & \cellcolor[rgb]{0.932,1.000,0.932} 58.3 \\
 & Human-scale & B4 classroom & \cellcolor[rgb]{0.780,1.000,0.780} 89.8 & \cellcolor[rgb]{0.869,1.000,0.869} 74.0 & \cellcolor[rgb]{0.796,1.000,0.796} 78.7 & \cellcolor[rgb]{0.931,1.000,0.931} 58.4 \\
 & Human-scale & B5 late & \cellcolor[rgb]{0.787,1.000,0.787} 88.5 & \cellcolor[rgb]{0.878,1.000,0.878} 72.3 & \cellcolor[rgb]{0.790,1.000,0.790} 79.5 & \cellcolor[rgb]{0.946,1.000,0.946} 56.6 \\
\addlinespace[1.5pt]
 & 100M tok & B1 balanced & \cellcolor[rgb]{0.788,1.000,0.788} 88.4 & \cellcolor[rgb]{0.792,1.000,0.792} 88.1 & \cellcolor[rgb]{0.826,1.000,0.826} 74.5 & \cellcolor[rgb]{0.886,1.000,0.886} 63.9 \\
 & 100M tok & B2 simult. & \cellcolor[rgb]{0.773,1.000,0.773} 91.1 & \cellcolor[rgb]{0.870,1.000,0.870} 73.9 & \cellcolor[rgb]{0.815,1.000,0.815} 76.1 & \cellcolor[rgb]{0.937,1.000,0.937} 57.6 \\
 & 100M tok & B3 sequential & \cellcolor[rgb]{0.764,1.000,0.764} 92.7 & \cellcolor[rgb]{0.870,1.000,0.870} 73.9 & \cellcolor[rgb]{0.812,1.000,0.812} 76.5 & \cellcolor[rgb]{0.939,1.000,0.939} 57.4 \\
 & 100M tok & B4 classroom & \cellcolor[rgb]{0.770,1.000,0.770} 91.6 & \cellcolor[rgb]{0.871,1.000,0.871} 73.6 & \cellcolor[rgb]{0.817,1.000,0.817} 75.8 & \cellcolor[rgb]{0.931,1.000,0.931} 58.4 \\
 & 100M tok & B5 late & \cellcolor[rgb]{0.849,1.000,0.849} 77.3 & \cellcolor[rgb]{0.810,1.000,0.810} 84.9 & \cellcolor[rgb]{0.906,1.000,0.906} 63.2 & \cellcolor[rgb]{0.895,1.000,0.895} 62.8 \\
\addlinespace[1.5pt]
 & 2B tok & B1 balanced & \cellcolor[rgb]{0.794,1.000,0.794} 87.3 & \cellcolor[rgb]{0.801,1.000,0.801} 86.5 & \cellcolor[rgb]{0.839,1.000,0.839} 72.7 & \cellcolor[rgb]{0.881,1.000,0.881} 64.4 \\
 & 2B tok & B2 simult. & \cellcolor[rgb]{0.785,1.000,0.785} 88.9 & \cellcolor[rgb]{0.882,1.000,0.882} 71.7 & \cellcolor[rgb]{0.810,1.000,0.810} 76.8 & \cellcolor[rgb]{0.951,1.000,0.951} 55.9 \\
 & 2B tok & B3 sequential & \cellcolor[rgb]{0.779,1.000,0.779} 89.9 & \cellcolor[rgb]{0.874,1.000,0.874} 73.1 & \cellcolor[rgb]{0.810,1.000,0.810} 76.8 & \cellcolor[rgb]{0.955,1.000,0.955} 55.5 \\
 & 2B tok & B4 classroom & \cellcolor[rgb]{0.783,1.000,0.783} 89.3 & \cellcolor[rgb]{0.870,1.000,0.870} 73.9 & \cellcolor[rgb]{0.807,1.000,0.807} 77.2 & \cellcolor[rgb]{0.947,1.000,0.947} 56.4 \\
 & 2B tok & B5 late & \cellcolor[rgb]{0.783,1.000,0.783} 89.3 & \cellcolor[rgb]{0.875,1.000,0.875} 73.0 & \cellcolor[rgb]{0.809,1.000,0.809} 76.9 & \cellcolor[rgb]{0.953,1.000,0.953} 55.7 \\
\addlinespace[1.5pt]
 & 24B tok & B1 balanced & \cellcolor[rgb]{0.754,1.000,0.754} 94.4 & \cellcolor[rgb]{0.758,1.000,0.758} \textbf{94.4} & \cellcolor[rgb]{0.799,1.000,0.799} 78.3 & \cellcolor[rgb]{0.812,1.000,0.812} \textbf{72.8} \\
 & 24B tok & B2 simult. & \cellcolor[rgb]{0.745,1.000,0.745} \textbf{96.0} & \cellcolor[rgb]{0.811,1.000,0.811} 84.7 & \cellcolor[rgb]{0.775,1.000,0.775} \textbf{81.6} & \cellcolor[rgb]{0.870,1.000,0.870} 65.8 \\
 & 24B tok & B3 sequential & \cellcolor[rgb]{0.748,1.000,0.748} 95.5 & \cellcolor[rgb]{0.822,1.000,0.822} 82.6 & \cellcolor[rgb]{0.779,1.000,0.779} 81.1 & \cellcolor[rgb]{0.881,1.000,0.881} 64.4 \\
\midrule
\texttt{zho--eng} & Human-scale & Mono & \textbf{-} & \cellcolor[rgb]{0.936,1.000,0.936} 61.7 & \cellcolor[rgb]{0.825,1.000,0.825} 74.6 & \cellcolor[rgb]{0.988,1.000,0.988} 51.5 \\
 & Human-scale & B1 balanced & \textbf{-} & \cellcolor[rgb]{0.855,1.000,0.855} 76.6 & \cellcolor[rgb]{0.855,1.000,0.855} 70.4 & \cellcolor[rgb]{0.868,1.000,0.868} 66.0 \\
 & Human-scale & B2 simult. & \textbf{-} & \cellcolor[rgb]{0.897,1.000,0.897} 68.8 & \cellcolor[rgb]{0.859,1.000,0.859} 69.8 & \cellcolor[rgb]{0.965,1.000,0.965} 54.2 \\
 & Human-scale & B3 sequential & \textbf{-} & \cellcolor[rgb]{0.889,1.000,0.889} 70.3 & \cellcolor[rgb]{0.857,1.000,0.857} 70.1 & \cellcolor[rgb]{0.961,1.000,0.961} 54.7 \\
 & Human-scale & B4 classroom & \textbf{-} & \cellcolor[rgb]{0.883,1.000,0.883} 71.4 & \cellcolor[rgb]{0.864,1.000,0.864} 69.1 & \cellcolor[rgb]{0.953,1.000,0.953} 55.7 \\
 & Human-scale & B5 late & \textbf{-} & \cellcolor[rgb]{0.893,1.000,0.893} 69.7 & \cellcolor[rgb]{0.848,1.000,0.848} 71.4 & \cellcolor[rgb]{0.960,1.000,0.960} 54.9 \\
\addlinespace[1.5pt]
 & 100M tok & B1 balanced & \textbf{-} & \cellcolor[rgb]{0.832,1.000,0.832} 80.8 & \cellcolor[rgb]{0.923,1.000,0.923} 60.9 & \cellcolor[rgb]{0.900,1.000,0.900} 62.2 \\
 & 100M tok & B2 simult. & \textbf{-} & \cellcolor[rgb]{0.815,1.000,0.815} 84.0 & \cellcolor[rgb]{0.921,1.000,0.921} 61.1 & \cellcolor[rgb]{0.891,1.000,0.891} 63.2 \\
 & 100M tok & B3 sequential & \textbf{-} & \cellcolor[rgb]{0.803,1.000,0.803} 86.2 & \cellcolor[rgb]{0.978,1.000,0.978} 53.1 & \cellcolor[rgb]{0.884,1.000,0.884} 64.1 \\
 & 100M tok & B4 classroom & \textbf{-} & \cellcolor[rgb]{0.894,1.000,0.894} 69.4 & \cellcolor[rgb]{0.910,1.000,0.910} 62.6 & \cellcolor[rgb]{0.939,1.000,0.939} 57.4 \\
 & 100M tok & B5 late & \textbf{-} & \cellcolor[rgb]{0.828,1.000,0.828} 81.6 & \cellcolor[rgb]{0.938,1.000,0.938} 58.7 & \cellcolor[rgb]{0.903,1.000,0.903} 61.8 \\
\addlinespace[1.5pt]
 & 2B tok & B1 balanced & \textbf{-} & \cellcolor[rgb]{0.799,1.000,0.799} 86.9 & \cellcolor[rgb]{0.904,1.000,0.904} 63.5 & \cellcolor[rgb]{0.889,1.000,0.889} 63.5 \\
 & 2B tok & B2 simult. & \textbf{-} & \cellcolor[rgb]{0.810,1.000,0.810} 84.8 & \cellcolor[rgb]{0.930,1.000,0.930} 59.8 & \cellcolor[rgb]{0.906,1.000,0.906} 61.4 \\
 & 2B tok & B3 sequential & \textbf{-} & \cellcolor[rgb]{0.815,1.000,0.815} 84.0 & \cellcolor[rgb]{0.955,1.000,0.955} 56.4 & \cellcolor[rgb]{0.893,1.000,0.893} 63.0 \\
 & 2B tok & B4 classroom & \textbf{-} & \cellcolor[rgb]{0.872,1.000,0.872} 73.4 & \cellcolor[rgb]{0.903,1.000,0.903} 63.7 & \cellcolor[rgb]{0.932,1.000,0.932} 58.2 \\
 & 2B tok & B5 late & \textbf{-} & \cellcolor[rgb]{0.841,1.000,0.841} 79.2 & \cellcolor[rgb]{0.933,1.000,0.933} 59.4 & \cellcolor[rgb]{0.911,1.000,0.911} 60.8 \\
 \addlinespace[1.5pt]
 & 24B tok (FW-24B) & B1 balanced & \textbf{-} & \cellcolor[rgb]{0.773,1.000,0.773} 91.7 & \cellcolor[rgb]{0.826,1.000,0.826} \textbf{75.9} & \cellcolor[rgb]{0.840,1.000,0.840} 70.9 \\
 & 24B tok (FW-24B) & B2 simult. & \textbf{-} & \cellcolor[rgb]{0.767,1.000,0.767} 92.9 & \cellcolor[rgb]{0.838,1.000,0.838} 73.7 & \cellcolor[rgb]{0.836,1.000,0.836} 71.6 \\
 & 24B tok (FW-24B) & B3 sequential & \textbf{-} & \cellcolor[rgb]{0.762,1.000,0.762} \textbf{93.9} & \cellcolor[rgb]{1.000,0.965,0.965} 55.9 & \cellcolor[rgb]{0.831,1.000,0.831} \textbf{72.7} \\
 & 24B tok (FW-24B) & B3 sequential {\scriptsize\textit{ewc}} & \textbf{-} & \cellcolor[rgb]{0.765,1.000,0.765} 93.2 & \cellcolor[rgb]{0.988,1.000,0.988} 57.5 & \cellcolor[rgb]{0.843,1.000,0.843} 70.4 \\
 & 24B tok (FW-24B) & B4 classroom & \textbf{-} & \cellcolor[rgb]{0.784,1.000,0.784} 89.6 & \cellcolor[rgb]{0.824,1.000,0.824} 76.2 & \cellcolor[rgb]{0.865,1.000,0.865} 67.3 \\
 & 24B tok (FW-24B) & B5 late & \textbf{-} & \cellcolor[rgb]{0.785,1.000,0.785} 89.5 & \cellcolor[rgb]{0.892,1.000,0.892} 67.5 & \cellcolor[rgb]{0.852,1.000,0.852} 69.1 \\
\bottomrule
\end{tabular}
\end{minipage}\hfill
\begin{minipage}[t]{0.49\linewidth}\centering
\begin{tabular}{ll l cc cc}
\toprule
& & & \multicolumn{2}{c}{\textbf{MultiBLiMP}} & \multicolumn{2}{c}{\textbf{MonoBLiMP}} \\
\cmidrule(lr){4-5}\cmidrule(lr){6-7}
\textbf{Pair} & \textbf{Data} & \textbf{Curr.} & \textbf{L1} & \textbf{Eng} & \textbf{L1} & \textbf{Eng} \\
\midrule
\texttt{eus--eng} & 100M tok & B1 balanced & \cellcolor[rgb]{0.738,1.000,0.738} \textbf{97.4} & \cellcolor[rgb]{0.806,1.000,0.806} \textbf{85.5} & \textbf{-} & \cellcolor[rgb]{0.900,1.000,0.900} \textbf{62.1} \\
 & 100M tok & B2 simult. & \cellcolor[rgb]{0.742,1.000,0.742} 96.7 & \cellcolor[rgb]{0.891,1.000,0.891} 69.9 & \textbf{-} & \cellcolor[rgb]{0.979,1.000,0.979} 52.6 \\
 & 100M tok & B3 sequential & \cellcolor[rgb]{0.738,1.000,0.738} \textbf{97.4} & \cellcolor[rgb]{0.889,1.000,0.889} 70.3 & \textbf{-} & \cellcolor[rgb]{0.967,1.000,0.967} 54.0 \\
 & 100M tok & B4 classroom & \cellcolor[rgb]{0.745,1.000,0.745} 96.0 & \cellcolor[rgb]{0.859,1.000,0.859} 75.8 & \textbf{-} & \cellcolor[rgb]{0.951,1.000,0.951} 56.0 \\
 & 100M tok & B5 late & \cellcolor[rgb]{0.762,1.000,0.762} 93.0 & \cellcolor[rgb]{0.829,1.000,0.829} 81.3 & \textbf{-} & \cellcolor[rgb]{0.912,1.000,0.912} 60.7 \\
\addlinespace[1.5pt]
 & 2B tok & B1 balanced & \cellcolor[rgb]{0.750,1.000,0.750} 95.2 & \cellcolor[rgb]{0.824,1.000,0.824} 82.2 & \textbf{-} & \cellcolor[rgb]{0.907,1.000,0.907} 61.3 \\
 & 2B tok & B2 simult. & \cellcolor[rgb]{0.742,1.000,0.742} 96.7 & \cellcolor[rgb]{0.886,1.000,0.886} 70.9 & \textbf{-} & \cellcolor[rgb]{0.985,1.000,0.985} 51.8 \\
 & 2B tok & B3 sequential & \cellcolor[rgb]{0.742,1.000,0.742} 96.7 & \cellcolor[rgb]{0.903,1.000,0.903} 67.8 & \textbf{-} & \cellcolor[rgb]{0.975,1.000,0.975} 53.0 \\
 & 2B tok & B4 classroom & \cellcolor[rgb]{0.742,1.000,0.742} 96.7 & \cellcolor[rgb]{0.884,1.000,0.884} 71.3 & \textbf{-} & \cellcolor[rgb]{0.972,1.000,0.972} 53.4 \\
 & 2B tok & B5 late & \cellcolor[rgb]{0.744,1.000,0.744} 96.3 & \cellcolor[rgb]{0.890,1.000,0.890} 70.1 & \textbf{-} & \cellcolor[rgb]{0.981,1.000,0.981} 52.3 \\
\midrule
\texttt{fil--eng} & 100M tok & B1 balanced & \textbf{-} & \cellcolor[rgb]{0.803,1.000,0.803} \textbf{86.1} & \textbf{-} & \cellcolor[rgb]{0.877,1.000,0.877} \textbf{65.0} \\
 & 100M tok & B2 simult. & \textbf{-} & \cellcolor[rgb]{0.862,1.000,0.862} 75.3 & \textbf{-} & \cellcolor[rgb]{0.926,1.000,0.926} 59.0 \\
 & 100M tok & B3 sequential & \textbf{-} & \cellcolor[rgb]{0.859,1.000,0.859} 75.8 & \textbf{-} & \cellcolor[rgb]{0.929,1.000,0.929} 58.6 \\
\midrule
\texttt{hin--eng} & 100M tok & B1 balanced & \cellcolor[rgb]{0.792,1.000,0.792} 87.6 & \cellcolor[rgb]{0.801,1.000,0.801} \textbf{86.4} & \textbf{-} & \cellcolor[rgb]{0.886,1.000,0.886} \textbf{63.8} \\
 & 100M tok & B2 simult. & \cellcolor[rgb]{0.770,1.000,0.770} 91.6 & \cellcolor[rgb]{0.907,1.000,0.907} 67.1 & \textbf{-} & \cellcolor[rgb]{0.962,1.000,0.962} 54.6 \\
 & 100M tok & B3 sequential & \cellcolor[rgb]{0.769,1.000,0.769} \textbf{91.7} & \cellcolor[rgb]{0.900,1.000,0.900} 68.4 & \textbf{-} & \cellcolor[rgb]{0.959,1.000,0.959} 55.0 \\
 & 100M tok & B4 classroom & \cellcolor[rgb]{0.792,1.000,0.792} 87.6 & \cellcolor[rgb]{0.859,1.000,0.859} 75.8 & \textbf{-} & \cellcolor[rgb]{0.940,1.000,0.940} 57.3 \\
 & 100M tok & B5 late & \cellcolor[rgb]{0.831,1.000,0.831} 80.6 & \cellcolor[rgb]{0.837,1.000,0.837} 79.9 & \textbf{-} & \cellcolor[rgb]{0.900,1.000,0.900} 62.1 \\
\addlinespace[1.5pt]
 & 2B tok & B2 simult. & \cellcolor[rgb]{0.770,1.000,0.770} 91.6 & \cellcolor[rgb]{0.908,1.000,0.908} 66.9 & \textbf{-} & \cellcolor[rgb]{0.949,1.000,0.949} 56.2 \\
\midrule
\texttt{ita--eng} & 100M tok & B1 balanced & \cellcolor[rgb]{0.825,1.000,0.825} 81.7 & \cellcolor[rgb]{0.794,1.000,0.794} \textbf{87.8} & \textbf{-} & \cellcolor[rgb]{0.895,1.000,0.895} \textbf{62.7} \\
 & 100M tok & B2 simult. & \cellcolor[rgb]{0.807,1.000,0.807} \textbf{84.8} & \cellcolor[rgb]{0.889,1.000,0.889} 70.3 & \textbf{-} & \cellcolor[rgb]{0.970,1.000,0.970} 53.7 \\
 & 100M tok & B3 sequential & \cellcolor[rgb]{0.807,1.000,0.807} \textbf{84.8} & \cellcolor[rgb]{0.877,1.000,0.877} 72.6 & \textbf{-} & \cellcolor[rgb]{0.965,1.000,0.965} 54.3 \\
 & 100M tok & B4 classroom & \cellcolor[rgb]{0.837,1.000,0.837} 79.5 & \cellcolor[rgb]{0.872,1.000,0.872} 73.5 & \textbf{-} & \cellcolor[rgb]{0.950,1.000,0.950} 56.1 \\
 & 100M tok & B5 late & \cellcolor[rgb]{0.907,1.000,0.907} 66.8 & \cellcolor[rgb]{0.822,1.000,0.822} 82.6 & \textbf{-} & \cellcolor[rgb]{0.905,1.000,0.905} 61.6 \\
\midrule
\texttt{pol--eng} & 100M tok & B1 balanced & \cellcolor[rgb]{0.846,1.000,0.846} 77.8 & \cellcolor[rgb]{0.800,1.000,0.800} \textbf{86.6} & \textbf{-} & \cellcolor[rgb]{0.900,1.000,0.900} \textbf{62.2} \\
 & 100M tok & B2 simult. & \cellcolor[rgb]{0.820,1.000,0.820} \textbf{82.5} & \cellcolor[rgb]{0.886,1.000,0.886} 70.9 & \textbf{-} & \cellcolor[rgb]{0.972,1.000,0.972} 53.4 \\
 & 100M tok & B3 sequential & \cellcolor[rgb]{0.827,1.000,0.827} 81.3 & \cellcolor[rgb]{0.896,1.000,0.896} 69.1 & \textbf{-} & \cellcolor[rgb]{0.960,1.000,0.960} 54.8 \\
\midrule
\texttt{rus--eng} & 100M tok & B1 balanced & \cellcolor[rgb]{0.848,1.000,0.848} 77.5 & \cellcolor[rgb]{0.809,1.000,0.809} \textbf{85.1} & \textbf{-} & \cellcolor[rgb]{0.910,1.000,0.910} \textbf{60.9} \\
 & 100M tok & B2 simult. & \cellcolor[rgb]{0.827,1.000,0.827} 81.3 & \cellcolor[rgb]{0.899,1.000,0.899} 68.6 & \textbf{-} & \cellcolor[rgb]{0.970,1.000,0.970} 53.7 \\
 & 100M tok & B3 sequential & \cellcolor[rgb]{0.822,1.000,0.822} \textbf{82.1} & \cellcolor[rgb]{0.890,1.000,0.890} 70.1 & \textbf{-} & \cellcolor[rgb]{0.972,1.000,0.972} 53.4 \\
 & 100M tok & B4 classroom & \cellcolor[rgb]{0.847,1.000,0.847} 77.7 & \cellcolor[rgb]{0.854,1.000,0.854} 76.8 & \textbf{-} & \cellcolor[rgb]{0.951,1.000,0.951} 55.9 \\
 & 100M tok & B5 late & \cellcolor[rgb]{0.910,1.000,0.910} 66.2 & \cellcolor[rgb]{0.827,1.000,0.827} 81.8 & \textbf{-} & \cellcolor[rgb]{0.918,1.000,0.918} 60.0 \\
\midrule
\texttt{tur--eng} & 100M tok & B1 balanced & \cellcolor[rgb]{0.853,1.000,0.853} 76.6 & \cellcolor[rgb]{0.812,1.000,0.812} \textbf{84.4} & \cellcolor[rgb]{0.825,1.000,0.825} 74.6 & \cellcolor[rgb]{0.909,1.000,0.909} \textbf{61.1} \\
 & 100M tok & B2 simult. & \cellcolor[rgb]{0.827,1.000,0.827} \textbf{81.2} & \cellcolor[rgb]{0.885,1.000,0.885} 71.0 & \cellcolor[rgb]{0.776,1.000,0.776} \textbf{81.5} & \cellcolor[rgb]{0.973,1.000,0.973} 53.3 \\
 & 100M tok & B3 sequential & \cellcolor[rgb]{0.831,1.000,0.831} 80.6 & \cellcolor[rgb]{0.893,1.000,0.893} 69.7 & \cellcolor[rgb]{0.781,1.000,0.781} 80.8 & \cellcolor[rgb]{0.973,1.000,0.973} 53.3 \\
 & 100M tok & B4 classroom & \cellcolor[rgb]{0.876,1.000,0.876} 72.5 & \cellcolor[rgb]{0.871,1.000,0.871} 73.6 & \cellcolor[rgb]{0.836,1.000,0.836} 73.1 & \cellcolor[rgb]{0.965,1.000,0.965} 54.2 \\
 & 100M tok & B5 late & \cellcolor[rgb]{0.877,1.000,0.877} 72.2 & \cellcolor[rgb]{0.829,1.000,0.829} 81.3 & \cellcolor[rgb]{0.918,1.000,0.918} 61.5 & \cellcolor[rgb]{0.909,1.000,0.909} 61.0 \\
\midrule
\texttt{eng--deu} & 24B tok & B1 balanced & \cellcolor[rgb]{0.759,1.000,0.759} \textbf{93.6} & \cellcolor[rgb]{0.762,1.000,0.762} \textbf{93.6} & \textbf{-} & \cellcolor[rgb]{0.822,1.000,0.822} \textbf{71.6} \\
\midrule
\texttt{eng--nld} & 24B tok & B1 balanced & \cellcolor[rgb]{0.759,1.000,0.759} 93.5 & \cellcolor[rgb]{0.763,1.000,0.763} 93.5 & \textbf{-} & \cellcolor[rgb]{0.817,1.000,0.817} 72.2 \\
 & 24B tok & B2 simult. & \cellcolor[rgb]{0.748,1.000,0.748} \textbf{95.5} & \cellcolor[rgb]{0.752,1.000,0.752} \textbf{95.5} & \textbf{-} & \cellcolor[rgb]{0.801,1.000,0.801} \textbf{74.2} \\
 & 24B tok & B3 sequential & \cellcolor[rgb]{0.757,1.000,0.757} 94.0 & \cellcolor[rgb]{0.760,1.000,0.760} 94.0 & \textbf{-} & \cellcolor[rgb]{0.802,1.000,0.802} 74.1 \\
\bottomrule
\end{tabular}
\end{minipage}
\caption{Accuracy of \beetle{} models across curricula and L1s and data exposure (100, 2B, 24B) on L1 and L2 MultiBLiMP \citep{jumelet2025multiblimp} and Monolingual BLiMPs (\texttt{English:} BLiMP \citep{warstadt2020blimp}); \texttt{Dutch:} BLiMP-NL \citep{suijkerbuijk-etal-2025-blimp-nl}; \texttt{Chinese:} ZhoBLiMP \citep{liu-etal-2025-zhoblimp}}
\label{tab:beetle_main}
\end{table*}

\begin{table}[!ht]
\centering\scriptsize
\setlength{\tabcolsep}{3pt}
\begin{tabular}{ll l cc cc}
\toprule
& & & \multicolumn{2}{c}{\textbf{MultiBLiMP}} & \multicolumn{2}{c}{\textbf{MonoBLiMP}} \\
\cmidrule(lr){4-5}\cmidrule(lr){6-7}
\textbf{Model} & \textbf{Par.} & \textbf{Cohort} & \textbf{L1} & \textbf{Eng} & \textbf{L1} & \textbf{Eng} \\
\midrule
\multirow{3}{*}{BLOOM-560M} & \multirow{3}{*}{0.56B} & Dutch & \cellcolor[rgb]{0.946,1.000,0.946} 59.8 & \cellcolor[rgb]{0.827,1.000,0.827} 81.7 & \cellcolor[rgb]{0.982,1.000,0.982} 52.6 & \cellcolor[rgb]{0.863,1.000,0.863} 66.7 \\
 &  & German & \cellcolor[rgb]{0.912,1.000,0.912} 65.9 & \cellcolor[rgb]{0.827,1.000,0.827} 81.7 &  & \cellcolor[rgb]{0.863,1.000,0.863} 66.7 \\
 &  & Chinese &  & \cellcolor[rgb]{0.827,1.000,0.827} 81.7 & \cellcolor[rgb]{0.840,1.000,0.840} 72.5 & \cellcolor[rgb]{0.863,1.000,0.863} 66.7 \\
\addlinespace[2pt]
\multirow{3}{*}{BLOOM-1.1B} & \multirow{3}{*}{1.1B} & Dutch & \cellcolor[rgb]{0.928,1.000,0.928} 63.0 & \cellcolor[rgb]{0.768,1.000,0.768} 92.5 & \cellcolor[rgb]{1.000,0.730,0.730} 48.7 & \cellcolor[rgb]{0.780,1.000,0.780} 76.7 \\
 &  & German & \cellcolor[rgb]{0.853,1.000,0.853} 76.5 & \cellcolor[rgb]{0.768,1.000,0.768} 92.5 &  & \cellcolor[rgb]{0.780,1.000,0.780} 76.7 \\
 &  & Chinese &  & \cellcolor[rgb]{0.768,1.000,0.768} 92.5 & \cellcolor[rgb]{0.758,1.000,0.758} 84.1 & \cellcolor[rgb]{0.780,1.000,0.780} 76.7 \\
\addlinespace[2pt]
\multirow{3}{*}{BLOOM-1.7B} & \multirow{3}{*}{1.7B} & Dutch & \cellcolor[rgb]{0.909,1.000,0.909} 66.4 & \cellcolor[rgb]{0.765,1.000,0.765} 93.0 & \cellcolor[rgb]{1.000,0.730,0.730} 48.7 & \cellcolor[rgb]{0.783,1.000,0.783} 76.4 \\
 &  & German & \cellcolor[rgb]{0.861,1.000,0.861} 75.2 & \cellcolor[rgb]{0.765,1.000,0.765} 93.0 &  & \cellcolor[rgb]{0.783,1.000,0.783} 76.4 \\
 &  & Chinese &  & \cellcolor[rgb]{0.765,1.000,0.765} 93.0 & \cellcolor[rgb]{0.753,1.000,0.753} 84.8 & \cellcolor[rgb]{0.783,1.000,0.783} 76.4 \\
\addlinespace[2pt]
\multirow{3}{*}{BLOOM-3B} & \multirow{3}{*}{3B} & Dutch & \cellcolor[rgb]{0.910,1.000,0.910} 66.3 & \cellcolor[rgb]{0.756,1.000,0.756} 94.8 & \cellcolor[rgb]{0.992,1.000,0.992} 51.1 & \cellcolor[rgb]{0.801,1.000,0.801} 74.2 \\
 &  & German & \cellcolor[rgb]{0.830,1.000,0.830} 80.7 & \cellcolor[rgb]{0.756,1.000,0.756} 94.8 &  & \cellcolor[rgb]{0.801,1.000,0.801} 74.2 \\
 &  & Chinese &  & \cellcolor[rgb]{0.756,1.000,0.756} 94.8 & \cellcolor[rgb]{0.758,1.000,0.758} 84.0 & \cellcolor[rgb]{0.801,1.000,0.801} 74.2 \\
\addlinespace[2pt]
\multirow{3}{*}{BLOOM-7.1B} & \multirow{3}{*}{7.1B} & Dutch & \cellcolor[rgb]{0.923,1.000,0.923} 64.0 & \cellcolor[rgb]{0.760,1.000,0.760} 94.0 & \cellcolor[rgb]{0.961,1.000,0.961} 55.5 & \cellcolor[rgb]{0.774,1.000,0.774} 77.5 \\
 &  & German & \cellcolor[rgb]{0.822,1.000,0.822} 82.2 & \cellcolor[rgb]{0.760,1.000,0.760} 94.0 &  & \cellcolor[rgb]{0.774,1.000,0.774} 77.5 \\
 &  & Chinese &  & \cellcolor[rgb]{0.760,1.000,0.760} 94.0 & \cellcolor[rgb]{0.756,1.000,0.756} 84.3 & \cellcolor[rgb]{0.774,1.000,0.774} 77.5 \\
\addlinespace[2pt]
\multirow{3}{*}{XGLM-4.5B} & \multirow{3}{*}{4.5B} & Dutch & \cellcolor[rgb]{0.741,1.000,0.741} 96.9 & \cellcolor[rgb]{0.734,1.000,0.734} 98.7 & \cellcolor[rgb]{0.761,1.000,0.761} 83.6 & \cellcolor[rgb]{0.746,1.000,0.746} 80.9 \\
 &  & German & \cellcolor[rgb]{0.730,1.000,0.730} 98.8 & \cellcolor[rgb]{0.734,1.000,0.734} 98.7 &  & \cellcolor[rgb]{0.746,1.000,0.746} 80.9 \\
 &  & Chinese &  & \cellcolor[rgb]{0.734,1.000,0.734} 98.7 & \cellcolor[rgb]{0.754,1.000,0.754} 84.6 & \cellcolor[rgb]{0.746,1.000,0.746} 80.9 \\
\addlinespace[2pt]
\multirow{3}{*}{EuroLLM-9B} & \multirow{3}{*}{9B} & Dutch & \cellcolor[rgb]{0.854,1.000,0.854} 76.4 & \cellcolor[rgb]{0.812,1.000,0.812} 84.4 & \cellcolor[rgb]{0.730,1.000,0.730} 88.0 & \cellcolor[rgb]{0.756,1.000,0.756} 79.6 \\
 &  & German & \cellcolor[rgb]{0.849,1.000,0.849} 77.3 & \cellcolor[rgb]{0.812,1.000,0.812} 84.4 &  & \cellcolor[rgb]{0.756,1.000,0.756} 79.6 \\
 &  & Chinese &  & \cellcolor[rgb]{0.812,1.000,0.812} 84.4 & \cellcolor[rgb]{0.762,1.000,0.762} 83.5 & \cellcolor[rgb]{0.756,1.000,0.756} 79.6 \\
\addlinespace[2pt]
\multirow{3}{*}{Gemma-3-270M} & \multirow{3}{*}{0.27B} & Dutch & \cellcolor[rgb]{0.814,1.000,0.814} 83.7 & \cellcolor[rgb]{0.744,1.000,0.744} 97.0 & \cellcolor[rgb]{0.859,1.000,0.859} 69.9 & \cellcolor[rgb]{0.763,1.000,0.763} 78.8 \\
 &  & German & \cellcolor[rgb]{0.776,1.000,0.776} 90.4 & \cellcolor[rgb]{0.744,1.000,0.744} 97.0 &  & \cellcolor[rgb]{0.763,1.000,0.763} 78.8 \\
 &  & Chinese &  & \cellcolor[rgb]{0.744,1.000,0.744} 97.0 & \cellcolor[rgb]{0.782,1.000,0.782} 80.7 & \cellcolor[rgb]{0.763,1.000,0.763} 78.8 \\
\addlinespace[2pt]
\multirow{3}{*}{Gemma-2-2B} & \multirow{3}{*}{2B} & Dutch & \cellcolor[rgb]{0.769,1.000,0.769} 91.7 & \cellcolor[rgb]{0.734,1.000,0.734} 98.7 & \cellcolor[rgb]{0.783,1.000,0.783} 80.5 & \cellcolor[rgb]{0.782,1.000,0.782} 76.5 \\
 &  & German & \cellcolor[rgb]{0.748,1.000,0.748} 95.6 & \cellcolor[rgb]{0.734,1.000,0.734} 98.7 &  & \cellcolor[rgb]{0.782,1.000,0.782} 76.5 \\
 &  & Chinese &  & \cellcolor[rgb]{0.734,1.000,0.734} 98.7 & \cellcolor[rgb]{0.770,1.000,0.770} 82.3 & \cellcolor[rgb]{0.782,1.000,0.782} 76.5 \\
\addlinespace[2pt]
\multirow{3}{*}{Gemma-2-9B} & \multirow{3}{*}{9B} & Dutch & \cellcolor[rgb]{0.772,1.000,0.772} 91.2 & \cellcolor[rgb]{0.734,1.000,0.734} 98.8 & \cellcolor[rgb]{0.756,1.000,0.756} 84.4 & \cellcolor[rgb]{0.775,1.000,0.775} 77.3 \\
 &  & German & \cellcolor[rgb]{0.743,1.000,0.743} 96.4 & \cellcolor[rgb]{0.734,1.000,0.734} 98.8 &  & \cellcolor[rgb]{0.775,1.000,0.775} 77.3 \\
 &  & Chinese &  & \cellcolor[rgb]{0.734,1.000,0.734} 98.8 & \cellcolor[rgb]{0.766,1.000,0.766} 82.9 & \cellcolor[rgb]{0.775,1.000,0.775} 77.3 \\
\addlinespace[2pt]
\multirow{3}{*}{Gemma-2-27B} & \multirow{3}{*}{27B} & Dutch & \cellcolor[rgb]{0.842,1.000,0.842} 78.6 & \cellcolor[rgb]{0.822,1.000,0.822} 82.6 & \cellcolor[rgb]{0.829,1.000,0.829} 74.0 & \cellcolor[rgb]{0.830,1.000,0.830} 70.7 \\
 &  & German & \cellcolor[rgb]{0.828,1.000,0.828} 81.1 & \cellcolor[rgb]{0.822,1.000,0.822} 82.6 &  & \cellcolor[rgb]{0.830,1.000,0.830} 70.7 \\
 &  & Chinese &  & \cellcolor[rgb]{0.822,1.000,0.822} 82.6 & \cellcolor[rgb]{0.847,1.000,0.847} 71.5 & \cellcolor[rgb]{0.830,1.000,0.830} 70.7 \\
\addlinespace[2pt]
\multirow{3}{*}{Qwen3-0.6B} & \multirow{3}{*}{0.6B} & Dutch & \cellcolor[rgb]{0.826,1.000,0.826} 81.5 & \cellcolor[rgb]{0.744,1.000,0.744} 97.0 & \cellcolor[rgb]{0.846,1.000,0.846} 71.7 & \cellcolor[rgb]{0.740,1.000,0.740} 81.6 \\
 &  & German & \cellcolor[rgb]{0.764,1.000,0.764} 92.6 & \cellcolor[rgb]{0.744,1.000,0.744} 97.0 &  & \cellcolor[rgb]{0.740,1.000,0.740} 81.6 \\
 &  & Chinese &  & \cellcolor[rgb]{0.744,1.000,0.744} 97.0 & \cellcolor[rgb]{0.773,1.000,0.773} 82.0 & \cellcolor[rgb]{0.740,1.000,0.740} 81.6 \\
\addlinespace[2pt]
\multirow{3}{*}{Qwen3-4B} & \multirow{3}{*}{4B} & Dutch & \cellcolor[rgb]{0.757,1.000,0.757} 93.9 & \cellcolor[rgb]{0.730,1.000,0.730} 99.5 & \cellcolor[rgb]{0.768,1.000,0.768} 82.6 & \cellcolor[rgb]{0.730,1.000,0.730} 82.8 \\
 &  & German & \cellcolor[rgb]{0.739,1.000,0.739} 97.2 & \cellcolor[rgb]{0.730,1.000,0.730} 99.5 &  & \cellcolor[rgb]{0.730,1.000,0.730} 82.8 \\
 &  & Chinese &  & \cellcolor[rgb]{0.730,1.000,0.730} 99.5 & \cellcolor[rgb]{0.745,1.000,0.745} 85.9 & \cellcolor[rgb]{0.730,1.000,0.730} 82.8 \\
\addlinespace[2pt]
\multirow{3}{*}{Qwen3-8B} & \multirow{3}{*}{8B} & Dutch & \cellcolor[rgb]{0.753,1.000,0.753} 94.6 & \cellcolor[rgb]{0.734,1.000,0.734} 98.7 & \cellcolor[rgb]{0.772,1.000,0.772} 82.1 & \cellcolor[rgb]{0.732,1.000,0.732} 82.5 \\
 &  & German & \cellcolor[rgb]{0.732,1.000,0.732} 98.5 & \cellcolor[rgb]{0.734,1.000,0.734} 98.7 &  & \cellcolor[rgb]{0.732,1.000,0.732} 82.5 \\
 &  & Chinese &  & \cellcolor[rgb]{0.734,1.000,0.734} 98.7 & \cellcolor[rgb]{0.745,1.000,0.745} 85.9 & \cellcolor[rgb]{0.732,1.000,0.732} 82.5 \\
\addlinespace[2pt]
\multirow{3}{*}{Llama-3.2-1B} & \multirow{3}{*}{1B} & Dutch & \cellcolor[rgb]{0.774,1.000,0.774} 90.8 & \cellcolor[rgb]{0.733,1.000,0.733} 99.0 & \cellcolor[rgb]{0.815,1.000,0.815} 76.0 & \cellcolor[rgb]{0.738,1.000,0.738} 81.8 \\
 &  & German & \cellcolor[rgb]{0.745,1.000,0.745} 96.1 & \cellcolor[rgb]{0.733,1.000,0.733} 99.0 &  & \cellcolor[rgb]{0.738,1.000,0.738} 81.8 \\
 &  & Chinese &  & \cellcolor[rgb]{0.733,1.000,0.733} 99.0 & \cellcolor[rgb]{0.766,1.000,0.766} 82.9 & \cellcolor[rgb]{0.738,1.000,0.738} 81.8 \\
\addlinespace[2pt]
\multirow{3}{*}{Llama-3.1-8B} & \multirow{3}{*}{8B} & Dutch & \cellcolor[rgb]{0.745,1.000,0.745} 96.1 & \cellcolor[rgb]{0.732,1.000,0.732} 99.2 & \cellcolor[rgb]{0.751,1.000,0.751} 85.1 & \cellcolor[rgb]{0.743,1.000,0.743} 81.2 \\
 &  & German & \cellcolor[rgb]{0.734,1.000,0.734} 98.0 & \cellcolor[rgb]{0.732,1.000,0.732} 99.2 &  & \cellcolor[rgb]{0.743,1.000,0.743} 81.2 \\
 &  & Chinese &  & \cellcolor[rgb]{0.732,1.000,0.732} 99.2 & \cellcolor[rgb]{0.755,1.000,0.755} 84.5 & \cellcolor[rgb]{0.743,1.000,0.743} 81.2 \\
\bottomrule
\end{tabular}
\caption{Frontier multilingual-LLM baselines. \emph{Par.}\ is the
parameter count; \emph{Cohort} is the evaluation language.}
\label{tab:baselines}
\end{table}

\clearpage

\end{document}